\documentclass[11pt]{article}

\usepackage[preprint]{acl}

\usepackage{times}
\usepackage{latexsym}
\usepackage{booktabs}
\usepackage{multirow}
\usepackage{tabularx}
\usepackage{makecell}
\usepackage{booktabs}
\usepackage{rotating}
\usepackage{array}
\usepackage{xcolor}
\usepackage{subcaption}
\newcommand{\posdrop}[1]{\textcolor{green!45!black}{#1}}
\newcommand{\negdrop}[1]{\textcolor{red!70!black}{#1}}

\usepackage[T1]{fontenc}

\usepackage[utf8]{inputenc}

\usepackage{microtype}

\usepackage{inconsolata}

\usepackage{graphicx}
\usepackage{colortbl}

\title{Interpretable Predictability-Based AI Text Detection: A Replication Study}

\author{
Adam Skurla$^{1,2}$,
Dominik Macko$^{2}$,
Jakub Simko$^{2}$ \\
  $^{1}$ Faculty of Information Technology, Brno University of Technology, Brno, Czechia\\
  $^{2}$ Kempelen Institute of Intelligent Technologies, Bratislava, Slovakia\\
\texttt{\{adam.skurla, dominik.macko, jakub.simko\}@kinit.sk}
}

\begin{document}
\maketitle
\begin{abstract}
This paper replicates and extends the system used in the AuTexTification shared task for authorship attribution of machine-generated texts. Exact replication was not possible because of differences in data splits, model availability, and implementation details, which we document as a case study in reproducibility. We tested newer multilingual language models (mDeBERTa-v3-base, Qwen, mGPT) and added 26 document-level stylometric features, using ablation, permutation importance, and SHAP analysis to assess feature influence. A single shared configuration was applied to both English and Spanish across Subtask~1 and Subtask~2. Averaged over three random seeds, the shared multilingual configuration performs comparably to or better than the language-specific baseline, with the clearest gains on model attribution (Subtask~2). The additional stylometric features yield small improvements, led by lexical diversity, but their contribution falls within seed variance once predictability-based probabilities are included, which remain the dominant signal. The study also shows that clear documentation is important for reliable replication and fair comparison of systems.
\end{abstract}

\section{Introduction}\label{sec:intro}

The rapid development of large language models (LLMs), such as ChatGPT \cite{openai2023gpt4} or LLaMA \cite{grattafiori2024llama}, has significantly impacted the field of text generation. These models are capable of producing fluent, stylistically convincing texts across multiple languages and domains at remarkable speed. Their outputs are often indistinguishable from human-written texts at first glance. As AI-generated texts become more prevalent in sectors like education and journalism, it raises important concerns, emphasizing the need for reliable automated tools to detect and attribute machine-generated texts (MGTs).

As the quality of generated texts improves, research has evolved from simple binary detection (human vs. AI) to the more complex task of model attribution. The goal is no longer just to determine whether a text was generated, but also to identify the specific model responsible. This task is more challenging because modern language models often generate outputs that are strikingly similar, with differences typically manifesting in subtle stylistic, lexical, or probabilistic features. At the same time, many recent approaches rely on fine-tuned neural models that behave largely as black boxes, which motivates the use of more interpretable methods for attribution \cite{luo2024understanding}.

In response to these challenges, several approaches have been proposed. One such approach of \citet{przybyla2023ve} was introduced in the AuTexTification 2023 shared task~\citep{sarvazyan2023overview}, where a hybrid system combined predictability features, derived from generative language models, with features from a fine-tuned language model (FLM) and additional linguistic features such as word frequency and grammatical correctness. This approach achieved competitive results in both binary detection and model attribution.

Yet competitive results on a shared task leaderboard do not, on their own, constitute robust scientific evidence. \textbf{Replication studies are a cornerstone of the scientific process}, enabling independent verification of empirical claims and exposing the limits of their generalizability. This concern is well-documented in NLP, where empiricism in computational linguistics still largely remains a matter of faith \citep{wieling-etal-2018-squib}, and influential replication work has repeatedly shown that reported advances are more fragile than they appear. \citet{melis2018on} demonstrated this forcefully, finding that standard LSTM language models, when properly tuned, match or outperform architectures that had been presented as superior, suggesting that many claimed advances reflect undertuned baselines rather than genuine progress. \citet{arvan-etal-2022-reproducibility-computational} showed that even the availability of source code is insufficient for reliable reproduction, achieving a replication success rate of only 25\% across a set of NLP papers with open code. The community has begun to address this problem through explicit inclusion of reproducing previous results (or failing to do so) in call for papers of top NLP venues (e.g., ACL or EMNLP), and dedicated initiatives such as the ReproNLP \citep{belz-etal-2025-2025}. Shared tasks offer a seemingly favorable setting for replication given their standardized data and evaluation protocols, yet system description papers produced in these settings remain rarely subjected to independent replication. This makes AuTexTification a suitable case study for testing whether a competitive hybrid detector remains reproducible, stable and interpretable under a controlled reimplementation.

\textbf{Our main goal is to replicate and extend the hybrid approach proposed for AuTexTification.}
We first reproduce the system of \citet{przybyla2023ve} to establish a comparable baseline and document the reproducibility constraints of the released setup. We then investigate whether the original language-specific components can be replaced by a shared multilingual configuration and whether additional document-level stylometric features provide useful and interpretable signals.

We address the following research questions:

\textbf{RQ1:} Can the language-specific encoder and predictability components of the original system be replaced by a shared multilingual configuration without reducing performance?

\textbf{RQ2:} Which document-level stylometric feature groups contribute useful signal, and what do they reveal about the model's decisions?

Unlike the original work, our focus is not only on classification performance, but also on identification of parts of the hybrid system responsible for the final predictions. 

Our main contributions are:

\textbf{(1)} A \textbf{systematic replication of the AuTexTification system}, identifying several factors that affect reproducibility of the reported results.

\textbf{(2)} A \textbf{unified multilingual configuration} based on recent language models, showing that language-specific components can be replaced while maintaining comparable or higher average performance.

\textbf{(3)} An \textbf{extended stylometric analysis} based on leave-one-group-out ablation, grouped permutation importance, and signed SHAP values.

To ensure reproducibility, the implementation code is publicly available.\footnote{\scriptsize\url{https://github.com/kinit-sk/AuTexTification-replication}}

\section{Related Work}

Early research in detection of machine-generated texts predominantly focused on \textbf{binary detection}, i.e., distinguishing between human- and machine-authored texts. Early milestones include Grover~\citep{zellers2019defending} and GLTR~\citep{gehrmann2019gltr}, which established that machine-generated texts are more predictable than human-written ones and can be detected accordingly.
DetectGPT~\citep{mitchell2023detectgpt} and its more efficient variant Fast-DetectGPT~\citep{bao2023fast} demonstrated that probabilistic features can be useful for binary detection tasks, although they remain constrained by their reliance on reference models and language-specific properties.

With the growing diversity of LLMs, the focus has shifted from binary detection \cite{spiegel-macko-2024-imgtb, hans2024spottingllmsbinocularszeroshot, su2023detectllm} to \textbf{multi-way attribution} \cite{mikros2023ai, alecakir2024groningen, wang2024m4gt, soto2024few}. The task is no longer only to detect MGT, but also to identify which model created it. 
Several approaches have been proposed to address this challenge. One approach is the use of stylometry \cite{przybyla2023ve, alecakir2024groningen, schaaff2023classificationhumanaigeneratedtexts}, where lexical and syntactic features of the text are analyzed. These methods are interpretable, but they often require high computational resources and do not work equally well across languages. Other approaches employ embedding-based \cite{wang2024trace, kuznetsov2024robust, kadhim2025adversarial} and contrastive methods \cite{guo2024detective, zhang2024nuanced}, leveraging representations from transformer models combined with contrastive learning to amplify distinctions between models.

Recent research on \textbf{multilingual} AI-text detection has moved beyond English-centric approaches, utilizing cross-lingual transformer architectures, such as XLM-RoBERTa \cite{conneau2019unsupervised} or mGPT \cite{shliazhko2024mgpt}. Multilingual evaluations \cite{macko-etal-2023-multitude, macko-etal-2025-multisocial, la-cava-etal-2026-authorship} indicate that fine-tuned multilingual transformer models outperform statistical and monolingual methods, achieving stronger cross-lingual generalization and highlighting the benefits of shared semantic representations for multilingual detection.

Tasks of MGT detection and attribution increasingly appear in the form of shared tasks, such as SemEval \cite{wang-etal-2024-semeval-2024}, AuTexTification \cite{chiruzzo2024overview}, or RuATD~\citep{Shamardina_2022}. Within AuTexTification, the system by \citet{przybyla2023ve} introduced an architecture that combined transformer-based embeddings, probabilistic features derived from language models, and traditional linguistic statistics. This approach improved performance, but it had some limitations. It used a simple feature combination, relied on older models, and covered only two languages. In our work, we therefore use this architecture as a baseline and propose an extended approach that directly addresses these shortcomings.

\section{Methodology} \label{chap:meth}

Our study first establishes a replication baseline for the original system and then addresses two research questions (see Section~\ref{sec:intro}): the impact of base language model selection and the contribution of the extended stylometric feature set. To provide the answers, we first reproduce the original experimental setup using the publicly available code\footnote{\tiny\url{https://github.com/piotrmp/autext}} (the original and extended system configurations are overviewed in Table~\ref{tab:configurations}). We then keep the configuration unchanged and replace the language models used to compute probabilistic features. Finally, we extend the stylometric feature set with additional indicators and evaluate their impact on model performance and interpretability. A schematic overview of the architecture and feature extraction pipeline is provided in Figure~\ref{fig:system_overview}.

\begin{table}[!b]
\centering
\setlength{\tabcolsep}{5pt}
\renewcommand{\arraystretch}{1.15}
\resizebox{\linewidth}{!}{%
\begin{tabular}{l | p{0.95\linewidth}}
\hline
\textbf{Configuration} & \textbf{Description} \\
\hline
FLM & Fine-tuned pre-trained encoder with [CLS] classification head. \\
\hline
Pred & BiLSTM over token-level probabilistic features. \\
\hline
Hybrid & FLM and Pred with concatenated outputs. \\
\hline
Hybrid+ & Hybrid extended with word frequency and grammar correctness features. \\
\hline
LingRF & Random Forest over document-level linguistic features. \\
\hline
LingRF + PredOut & LingRF concatenated with Pred output probabilities. \\
\hline
\hline
Ultrahybrid & Hybrid+ output probabilities + linguistic features into RF/XGB/MLP. \\
\hline
Hybrid\_flat & Statistically pooled token-level features + linguistic features into RF/XGB/MLP (no BiLSTM/FLM). \\
\hline
\end{tabular}%
}
\caption{Overview of system configurations (variants). The original configurations of \citet{przybyla2023ve} are provided in the upper part, the lower part contains the proposed additional configurations.}
\label{tab:configurations}
\end{table}

\textbf{Dataset}. We conduct our experiments on the AuTexTification 2023 dataset\footnote{\tiny\url{https://huggingface.co/datasets/symanto/autextification2023}}~\citep{sarvazyan2023overview}, following the original study. The shared task was organized in two languages, English and Spanish, and consisted of two subtasks.
The first subtask is a binary classification task (Human vs. AI), where the goal is to determine whether a text was written by a human or generated by a language model. The second subtask focuses on model attribution: all texts are machine-generated, and the objective is to identify which model produced each text.
During the competition, only limited information about the language models was provided. Participants were informed that six models were used, ranging from 2B to 175B parameters, and labeled A--F\footnote{our experiments were completely agnostic to known generator identities, which are BLOOM-1B1, BLOOM-3B, BLOOM-7B1, Babbage, Curie, and text-davinci-003}.
The dataset covers five genres: Legal, News, Reviews, Tweets, and Wiki. The training split includes Legal, Tweets, and Wiki, while the test split consists of News and Reviews. This setup evaluates \textbf{a challenging cross-domain generalization}, as the test domains are not seen during training. 
%
%
The dataset is balanced across classes and domains. In Subtask~1, the dataset includes about 16k samples for training and 10k samples for testing per each class in each language. In Subtask~2, there are about 3.6k samples for training and 900 samples for testing per each class (generator) in each language.
We follow the official train–test split for the shared task.

\textbf{Replication setup}. In this part, we focus on replicating the original system, primarily based on the information provided in the paper. We implemented the system configuration, feature set, and training procedure according to the available methodological description. A more detailed analysis of the original study is provided in Appendix~\ref{sec:appendix}. The authors also released the implementation of their best-performing system, which we used to verify implementation details and ensure consistency of the individual components. The objective was to preserve the implementation described in the paper. During replication, we retained the originally reported hyperparameters and evaluation setup, including the official train–test split.

\textbf{(RQ1) Base model variants}. In RQ1, we study whether the language-specific components of the original system can be replaced with a shared multilingual configuration. This question is motivated by \textbf{two limitations of the baseline}. First, the probabilistic features rely on GPT-2-style reference models, which provide uneven multilingual coverage. Second, the FLM component uses separate encoders for English and Spanish, making the system harder to maintain and extend to additional languages. A multilingual configuration would reduce language-specific engineering while preserving the same hybrid architecture. We therefore vary only the base models used in two parts of the pipeline: (i) the generative language models used to compute probabilistic, predictability-based features, and (ii) the encoder used to extract contextual representations in the FLM module. All other components remain unchanged, which isolates the effect of the selected base models. For probabilistic features, we compare three multilingual reference-model groups by approximate parameter scale and computational cost: \textit{(a) Small}: XGLM-564M, XGLM-1.7B, Qwen2.5-1.5B, and BLOOM-1.1B; \textit{(b) Medium}: mGPT, Llama-3.2-3B, Qwen2.5-1.5B, and BLOOM-1.7B; and \textit{(c) Large}: Qwen2.5-3B, Llama-3.2-3B, XGLM-2.9B, and BLOOM-1.7B. These groups cover different multilingual autoregressive model families while keeping feature extraction computationally feasible. The Large group further tests whether higher-capacity reference models provide stronger predictability signals.

For the FLM component, we compare the original language-specific encoder setup with two multilingual models: XLM-RoBERTa-base \cite{conneau2019unsupervised} and mDeBERTa-v3-base \cite{he2021debertav3}. XLM-RoBERTa is used as a common multilingual baseline. mDeBERTa-v3 is included because it showed the best overall results among multilingual fine-tuned detectors in the MULTITuDE benchmark~\citep{macko-etal-2023-multitude}. The aim is to test whether one shared encoder can be used for both English and Spanish without changing the rest of the system.

\textbf{(RQ2) Extension of stylometric features}. In RQ2, we extend the document-level stylometric representation of the original system. This is motivated by the assumption that MGTs may differ from human-written texts not only in token probabilities, but also in broader stylistic properties. Rather than proposing new stylometric descriptors, we test whether established document-level indicators provide complementary signal to neural and probabilistic features.

The original study included several linguistic feature groups, but did not provide a complete specification of the extracted features (Appendix~\ref{sec:appendix}). We therefore add 26 stylometric features covering seven categories commonly used in stylometric and authorship-analysis research: lexical diversity, word-level statistics, sentence-level structure, readability, repetition, punctuation, and functional markers. These categories are motivated by recent MGT detection work that uses linguistic-stylistic features to capture lexical richness, text statistics, readability, repetition, and punctuation patterns~\citep{darwinkel2024groningen, petukhova2024petkaz, sharma2024team}. Frequency-based measures are normalized by the number of tokens or sentences to reduce text-length effects. The full feature list is provided in Appendix~\ref{sec:app_features}. We compare the extended feature set against the reconstructed original feature set and analyze its contribution using leave-one-group-out ablation, grouped permutation importance, and signed SHAP values.

\textbf{Training and implementation details.} We follow the official train–test split, with 20\% of training data reserved for validation. For comparability, we use the same random seed (10) as the original study. All hyperparameters, early stopping, and Random Forest classifier settings follow the original study (see Appendix~\ref{sec:appendix}). Experiments were run on an NVIDIA A40 GPU (~400 GPU hours). We report performance using \textbf{Macro F1} as the primary evaluation metric (the official metric of the shared task).

\section{Replication Baseline}

When attempting to reproduce the system proposed by \citet {przybyla2023ve}, we obtained results that differed from those reported in the original publication. Based on these differences, we conducted a detailed analysis of the official repository referenced in the paper. Although the repository does not include implementations of all components (e.g., the linguistic features and the LingRF model are missing), its inspection allowed us to identify several factors that affect the reproducibility of the system. These factors can be grouped into three categories: (i) differences between the methodological description in the paper and the available implementation, (ii) external dependencies and model availability, and (iii) ambiguities in the implementation of linguistic features.

\textbf{Differences between the paper and the implementation}.  
The first group concerns differences between the experimental setup described in the paper and the publicly available implementation. While the paper reports a random 80/20 train–validation split, the available code uses a predefined topic-based split (fold-0 from the LDA split). This difference may result in a validation set with different properties and can therefore influence model selection. The paper also describes an early stopping strategy based on selecting the first epoch that reaches $f_{\max} - 0.01$. In the analyzed version of the code, however, this mechanism is not explicitly implemented; the model is trained for a fixed 20 epochs and predictions are stored for each epoch. It is therefore not fully clear how the final model was selected in the original experiments. We also acknowledge that the publicly available implementation may not exactly correspond to the final version used for the shared-task submission.

\textbf{Model availability for Spanish}.  
An important limitation of our reproduction concerns the availability of the language models used for Spanish. For extracting probabilistic features, the authors relied on \texttt{PlanTL-GOB-ES/gpt2-base-bne} and \texttt{PlanTL-GOB-ES/gpt2-large-bne}. At the time of our experiments (11/2025), these models were no longer available in their original form, preventing their direct use. In our baseline reproduction, we therefore replaced them with the closest available alternatives, namely \texttt{DeepESP/gpt2-spanish} and \texttt{datificate/gpt2-small-spanish}. Since these are different models with different training data and vocabularies, identical results cannot be expected, which may partly explain the differences observed for Spanish.

\textbf{Ambiguities in feature extraction}.  
The third group of factors relates to uncertainties in feature extraction. For grammatical features, the implementation compares tokens within a context window of $\pm 5$ positions, which is not specified in the paper and slightly alters their definition. Word-frequency features also depend on the tokenization and its alignment with the external frequency resource. For document-level aggregated linguistic features, the authors did not provide an implementation or an exact list of features (only their types are described), so we relied on the information available in the paper. The resulting feature representation depends on the version of the spaCy library and the aggregation of linguistic categories. The models \texttt{en\_core\_web\_sm} and \texttt{es\_core\_news\_sm} may produce annotations that differ from those at the time of publication, affecting the number of extracted features. We also observed small output differences between versions 4.X and 5.X of the Hugging Face Transformers library, likely due to internal implementation changes.

We also compared the reconstructed linguistic feature counts with those reported in the original paper. The counts closely match for English, whereas larger differences appear for Spanish, likely due to richer Spanish morphology and differences in spaCy versions and linguistic-category aggregation. A detailed comparison is provided in Appendix~\ref{sec:appendix}.

Despite the differences discussed above, we were able to reproduce the core experimental pipeline. In addition, we successfully ran the repository officially referenced by the authors, though we had to introduce several minor adjustments, including replacing unavailable Spanish language models and adding early stopping according to the criterion $f_{\max} - 0.01$. Table~\ref{tab:rq1_full} presents three sets of results: the scores reported in the original publication (Paper), the results obtained by running the publicly available code (GitHub), and the results of our reproduction (Ours).

\begin{table}[!b]
\centering
\setlength{\tabcolsep}{3pt}
\resizebox{0.75\columnwidth}{!}{%
\begin{tabular}{lllrrr}
\hline
\textbf{Subtask} & \textbf{Lang} & \textbf{Method}
& \textbf{Paper} & \textbf{GitHub} & \textbf{Ours} \\
\hline
\multirow{4}{*}{S1}
& \multirow{4}{*}{en}
& FLM                & n/a   & \textbf{0.584} & 0.564 \\
& & Pred             & n/a   & 0.850 & \textbf{0.887} \\
& & Hybrid       & 0.740 & 0.603 & \textbf{0.810} \\
& & Hybrid+ & \textbf{0.810} & 0.699 & 0.796 \\
\hline
\multirow{4}{*}{S1}
& \multirow{4}{*}{es}
& FLM                & n/a   & 0.550 & \textbf{0.551} \\
& & Pred             & n/a   & 0.740 & \textbf{0.764} \\
& & Hybrid       & 0.680 & 0.659 & \textbf{0.717} \\
& & Hybrid+  & 0.710 & 0.655 & \textbf{0.743} \\
\hline
\multirow{4}{*}{S2}
& \multirow{4}{*}{en}
& FLM                & n/a   & 0.562 & \textbf{0.573} \\
& & Pred             & n/a   & 0.441 & \textbf{0.478} \\
& & Hybrid       & 0.576 & 0.558 & \textbf{0.584} \\
& & Hybrid+ & 0.570 & 0.544 & \textbf{0.586} \\
\hline
\multirow{4}{*}{S2}
& \multirow{4}{*}{es}
& FLM                & n/a   & 0.559 & \textbf{0.583} \\
& & Pred             & n/a   & 0.340 & \textbf{0.421} \\
& & Hybrid       & \textbf{0.620} & 0.534 & 0.592 \\
& & Hybrid+ & \textbf{0.610} & 0.553 & 0.567 \\
\hline
\end{tabular}%
}
\caption{Test macro $F_1$ comparison of original reported results (Paper), results obtained by running the official code (GitHub), and results produced by our reproduction (Ours). Subtask~1 (S1) is binary human/machine detection; Subtask~2 (S2) is multiway generator attribution.}
\label{tab:rq1_full}
\end{table}

We observe strong performance of the model based solely on predictability-based features (Pred), particularly in the binary detection setting (S1), where it achieves the highest score among all variants. In this setting, adding additional features to the Pred model slightly decreases performance compared to the standalone variant, suggesting that predictability-based features alone already capture most of the relevant signal.

An exact numerical reproduction of the originally reported results was not achieved. Across settings where comparison with the paper was possible, the differences between the reported and our results range from 0.008 to 0.070 F1 points, with the largest gap observed for S1 EN (Hybrid). These deviations indicate that even relatively small differences in data splits, early stopping procedures, model availability, or feature extraction details can lead to measurable changes in performance. Our reproduced results are therefore used as the baseline reference in RQ1 and RQ2. \textbf{Our findings underline the importance of making the final experimental code fully available and ensuring consistency between the published methodological description and the released implementation in order to support reliable reproducibility.}

\section{RQ1: Impact of Base Language Models}

In this section, we analyze how the choice of base language models affects system performance. The system configuration, training procedure, and feature set remain unchanged; we only vary the generative models used to compute probabilistic features and the encoder in the FLM component. Our goal is to verify whether it is possible to create a unified multilingual configuration that can be used for both languages and both subtasks without language-specific modifications. We report results for the two strongest configurations from the original work, namely Hybrid and Hybrid+ (see Table~\ref{tab:configurations}). Unless stated otherwise, test-set results in this section are reported as the mean and standard deviation over three random seeds (10/11/12); development-set results used only for model selection are reported for a single seed.

\textbf{Impact of the encoder in the FLM component}.
The goal was to determine whether newer multilingual encoders provide measurable improvements over the original configuration. As mentioned in Section~\ref{chap:meth}, we compare multilingual encoders with the original language-specific encoder setup. Table~\ref{tab:rq2_flm_encoders} summarizes development results used for encoder selection and test results used for comparison with the reproduced baseline.

\begin{table}[!b]
\centering
\small
\setlength{\tabcolsep}{5pt}
\resizebox{\columnwidth}{!}{%
\begin{tabular}{llccccc}
\hline
\textbf{Config.} & \textbf{Encoder}
& \textbf{S1-en} & \textbf{S1-es}
& \textbf{S2-en} & \textbf{S2-es}
& \textbf{Avg.} \\
\hline
\multicolumn{7}{c}{\textit{Development set: encoder selection}} \\
\hline
\multirow{2}{*}{Hybrid}
& XLM-R       & 0.941 & 0.865 & 0.583 & 0.605 & 0.749 \\
& mDeBERTa-v3 & 0.952 & 0.936 & 0.597 & 0.624 & \textbf{0.777} \\
\addlinespace[2pt]
\multirow{2}{*}{Hybrid+}
& XLM-R       & 0.943 & 0.923 & 0.477 & 0.605 & 0.737 \\
& mDeBERTa-v3 & 0.950 & 0.937 & 0.594 & 0.625 & \textbf{0.776} \\
\hline
\multicolumn{7}{c}{\textit{Test set: Replication baseline reference}} \\
\hline
\multirow{2}{*}{Hybrid}
& Baseline    & 0.810 & 0.717 & 0.584 & 0.592 & \textbf{0.676} \\
& mDeBERTa-v3 & 0.708 & 0.672 & 0.604 & 0.617 & 0.650 \\
\addlinespace[2pt]
\multirow{2}{*}{Hybrid+}
& Baseline    & 0.796 & 0.743 & 0.586 & 0.567 & \textbf{0.673} \\
& mDeBERTa-v3 & 0.690 & 0.715 & 0.593 & 0.611 & 0.652 \\
\hline
\end{tabular}
}
\caption{Macro $F_1$ for encoder selection on the development set and comparison with the reproduced baseline on the test set. Test results shown here are single-seed (seed 10); multi-seed variance for the full multilingual configuration is reported in Table~\ref{tab:rq2_pred_groups} and Appendix~\ref{timecomplexity_baseline}.}
\label{tab:rq2_flm_encoders}
\end{table}

Development results select mDeBERTa-v3 over XLM-R for both Hybrid and Hybrid+ configurations. Compared with the reproduced baseline on the test set, the language-specific encoder setup remains stronger on average, mainly because of higher Subtask~1 performance. On Subtask~1 the difference between mDeBERTa-v3 and the original encoder falls within the seed variance reported below and should not be read as a robust degradation; on Subtask~2, however, mDeBERTa-v3 consistently improves both language settings while providing a single encoder shared across English and Spanish. We therefore use mDeBERTa-v3 in the following multilingual experiments, noting that its benefit is concentrated in the attribution subtask.

\textbf{Impact of generative models on probabilistic features}.
In the next step, we analyze how the choice of generative models used to compute predictability-based features affects the overall system performance. As in the previous experiment, we evaluate only the Hybrid and Hybrid+ configurations. The difference is that we now use the selected multilingual encoder mDeBERTa-v3. The experiment was conducted in several iterations; in this section, we report the final setting in which we compare three multilingual model groups (see Section~\ref{chap:meth}), denoted as Small, Medium, and Large. The goal is to determine which group provides the highest performance across languages and subtasks. These groups combine different model families and sizes and are intended as practical multilingual configurations rather than a controlled scaling study; we therefore do not isolate the effect of model capacity from that of model family.

The results are shown in Table~\ref{tab:rq2_pred_groups}. The Large group obtains the highest development average in both the Hybrid and Hybrid+ configurations. The Medium group is competitive in several settings, especially for Hybrid, while the Small group generally obtains lower average development performance. Based on development performance, we select the Large group for the final multilingual configuration.

\begin{table*}[!t]
\centering
\small
\setlength{\tabcolsep}{5pt}
\resizebox{1.5\columnwidth}{!}{%
\begin{tabular}{llccccc}
\hline
\textbf{Config.} & \textbf{Prob. group}
& \textbf{S1-en} & \textbf{S1-es}
& \textbf{S2-en} & \textbf{S2-es}
& \textbf{Avg.} \\
\hline
\multicolumn{7}{c}{\textit{Development set: probabilistic model selection}} \\
\hline
\multirow{3}{*}{Hybrid}
& Small  & 0.950 & 0.940 & 0.591 & 0.624 & 0.776 \\
& Medium & 0.945 & 0.953 & 0.606 & 0.644 & 0.787 \\
& Large  & 0.959 & 0.951 & 0.619 & 0.649 & \textbf{0.794} \\
\addlinespace[2pt]
\multirow{3}{*}{Hybrid+}
& Small  & 0.950 & 0.946 & 0.614 & 0.629 & 0.785 \\
& Medium & 0.947 & 0.953 & 0.597 & 0.636 & 0.783 \\
& Large  & 0.959 & 0.951 & 0.609 & 0.640 & \textbf{0.790} \\
\hline
\multicolumn{7}{c}{\textit{Test set: baseline vs. multilingual (mean\,$\pm$\,std, seeds 10/11/12)}} \\
\hline
\multirow{2}{*}{Hybrid}
& Baseline
& 0.788 {$\pm$ 0.034} & 0.721 {$\pm$ 0.006}
& 0.580 {$\pm$ 0.011} & 0.589 {$\pm$ 0.003} & 0.669 \\
& Large
& 0.828 {$\pm$ 0.023} & 0.747 {$\pm$ 0.044}
& \textbf{0.632} {$\pm$ 0.009} & \textbf{0.639} {$\pm$ 0.008} & \textbf{0.712} \\
\addlinespace[2pt]
\multirow{2}{*}{Hybrid+}
& Baseline
& 0.793 {$\pm$ 0.077} & 0.728 {$\pm$ 0.012}
& 0.588 {$\pm$ 0.002} & 0.573 {$\pm$ 0.011} & 0.671 \\
& Large
& 0.790 {$\pm$ 0.031} & 0.769 {$\pm$ 0.047}
& \textbf{0.620} {$\pm$ 0.017} & \textbf{0.637} {$\pm$ 0.002} & \textbf{0.704} \\
\hline
\end{tabular}
}
\caption{Macro $F_1$ for probabilistic model selection on the development set (single seed) and comparison with the reproduced baseline on the test set (mean\,$\pm$\,std over seeds 10/11/12). Bold test-set values mark settings where the multilingual Large configuration exceeds the baseline with non-overlapping $\pm1\sigma$ ranges.}
\label{tab:rq2_pred_groups}
\end{table*}

After selecting the Large group on the development split, we evaluate the resulting multilingual configuration on the official test set. This configuration combines the mDeBERTa-v3 encoder with the Large probabilistic model group and is referred to as the \textit{multilingual} configuration throughout the paper. Compared with the language-specific baseline, it provides a single solution for both languages and both subtasks. Averaged over three seeds, the multilingual configuration improves over the replicated baseline on both Subtask~2 settings, in both the Hybrid and Hybrid+ configurations, with non-overlapping $\pm1\sigma$ ranges. On Subtask~1 the two configurations are comparable: the mean is higher for the multilingual configuration in most settings, but the $\pm1\sigma$ ranges overlap, and for Hybrid+ on Subtask~1 English the difference is negligible. The gains are therefore concentrated in the attribution subtask, where subtle differences in token predictability appear to matter more. We interpret this as an association observed for the evaluated group composition rather than a controlled capacity effect, and we do not claim that larger reference models are intrinsically more stable. We further report per-seed variance and computational cost in Appendix~\ref{timecomplexity_baseline}, and additional architectural variants (Ultrahybrid and Hybrid\_flat) in Appendix~\ref{chap:new_models}. Overall, these results suggest that the multilingual configuration is a viable single-model replacement for the language-specific baseline, most clearly for attribution, while for binary detection it performs comparably rather than better.

\section{RQ2: Additional Stylometric Features} \label{chap:RQ3}

The original study states as one of its main limitations that the used features do not sufficiently capture deeper stylistic and structural characteristics of the text. Therefore, we analyze the contribution of newly introduced document-level stylometric features added to the original set of linguistic features. The goal is to examine whether extending the text representation improves classification performance and whether the new features provide additional information beyond the original setup. All test-set results in this section are reported as the mean and standard deviation over three random seeds (10/11/12).

We compare two feature sets. The first one uses the reconstructed original feature set from the replicated study. The second one extends this set with 26 additional document-level stylometric features described in Appendix~\ref{sec:app_features}. In both cases, the model configuration and training setup remain unchanged in order to isolate the effect of the extended feature set. We also analyze the new features using leave-one-group-out ablation, grouped permutation importance, and signed SHAP values.

\textbf{LingRF}. Table~\ref{tab:lingrf_rq2} presents the results for the LingRF model, which relies only on document-level linguistic features and does not use probabilistic outputs from the neural component.

\begin{table}[!t]
\centering
\resizebox{0.95\columnwidth}{!}{%
\begin{tabular}{llccc}
\hline
\textbf{Subtask} & \textbf{Lang} & \textbf{Baseline} & \textbf{Extended} & \textbf{$\Delta$} \\
\hline
S1 & en & 0.409 {$\pm$ 0.002} & 0.427 {$\pm$ 0.028} & +0.018 \\
S1 & es & 0.502 {$\pm$ 0.005} & 0.527 {$\pm$ 0.057} & +0.025 \\
S2 & en & 0.407 {$\pm$ 0.001} & 0.422 {$\pm$ 0.040} & +0.015 \\
S2 & es & 0.429 {$\pm$ 0.008} & 0.446 {$\pm$ 0.038} & +0.017 \\
\hline
\end{tabular}
}
\caption{Macro $F_1$ comparison of the original and extended feature sets for LingRF on the test set (mean\,$\pm$\,std over seeds 10/11/12).}
\label{tab:lingrf_rq2}
\end{table}

Adding the stylometric features raises the mean LingRF score in every language and subtask combination, by +0.015 to +0.025 macro $F_1$. However, these gains are small relative to seed variance: the $\pm1\sigma$ ranges of the baseline and extended feature sets overlap in all four settings, so with three runs the improvement cannot be clearly separated from seed noise. The extended representation therefore appears mildly beneficial for the standalone linguistic model, but we do not claim a statistically robust gain. The added variance of the extended set (larger standard deviations, e.g.\ $\pm.057$ for Spanish Subtask~1) also suggests that some of the new features behave unstably in the Random Forest.

\textbf{LingRF + PredOut}. We then analyze LingRF + PredOut, which combines document-level stylometric features with the probability outputs of the Pred component. Table~\ref{tab:lingrf_predout_rq2} reports three configurations: (i) \textit{Baseline}, the original probabilistic models with the original feature set, (ii) \textit{Old-prob}, the original probabilistic models with the extended feature set, and (iii) \textit{Multilingual}, the multilingual configuration from RQ1 with the extended feature set.

\begin{table}[!t]
\centering
\resizebox{\columnwidth}{!}{%
\begin{tabular}{llccc}
\hline
\textbf{Subtask} & \textbf{Lang} & \textbf{Baseline} & \textbf{Old-prob} & \textbf{Multilingual} \\
\hline
S1 & en & 0.875 {$\pm$ 0.010} & 0.877 {$\pm$ 0.009} & \textbf{0.889} {$\pm$ 0.021} \\
S1 & es & 0.770 {$\pm$ 0.006} & 0.779 {$\pm$ 0.007} & \textbf{0.867} {$\pm$ 0.010} \\
S2 & en & 0.493 {$\pm$ 0.018} & 0.501 {$\pm$ 0.031} & \textbf{0.557} {$\pm$ 0.030} \\
S2 & es & 0.453 {$\pm$ 0.016} & 0.463 {$\pm$ 0.040} & \textbf{0.542} {$\pm$ 0.015} \\
\hline
\end{tabular}%
}
\caption{Macro $F_1$ for LingRF+PredOut on the test set (mean\,$\pm$\,std over seeds 10/11/12). Bold marks settings where \textit{Multilingual} exceeds \textit{Baseline} with non-overlapping $\pm1\sigma$ ranges.}
\label{tab:lingrf_predout_rq2}
\end{table}

Two effects can be separated in Table~\ref{tab:lingrf_predout_rq2}. First, adding the extended features while keeping the original probabilistic models (\textit{Baseline}~$\rightarrow$~\textit{Old-prob}) yields only small gains of +0.002 to +0.010, with overlapping $\pm1\sigma$ ranges in all four settings; the stylometric features alone therefore add little once PredOut probabilities are present. Second, replacing the probabilistic models with the multilingual configuration (\textit{Old-prob}~$\rightarrow$~\textit{Multilingual}) produces substantially larger gains, most clearly on the attribution subtask, where \textit{Multilingual} exceeds \textit{Baseline} with non-overlapping $\pm1\sigma$ ranges in both languages (S2) and for Spanish Subtask~1. On English Subtask~1 the difference is within seed variance. The dominant signal in this combined model thus comes from the probabilistic configuration rather than from the added stylometric features. The comparison with the detector mdok~\citep{macko2025mdok} in Appendix~\ref{sec:comparison_sota} is included as a competitiveness reference rather than a matched benchmark (see Appendix~\ref{sec:comparison_sota} for the fine-tuning configuration).

\textbf{Feature-group diagnostics}. We analyse the extended representation using three methods: leave-one-group-out ablation, grouped permutation importance, and SHAP (SHapley Additive exPlanations) \cite{lundberg2017unified}. Ablation removes one stylometric group and retrains the RF. Permutation importance measures the performance drop after shuffling a feature group in the final multilingual model. SHAP is used to analyse feature importance and direction. Full results and grouped SHAP plots are provided in Appendix~\ref{sec:Abl_study} and~\ref{sec:app_shap}.

Table~\ref{tab:rq2_stylometric_diagnostics} summarizes the main ablation and permutation results. Positive values indicate a decrease in test macro $F_1$ after removing or permuting a feature group.
\begin{table}[!t]
\centering
\small
\setlength{\tabcolsep}{4pt}
\resizebox{0.9\columnwidth}{!}{%
\begin{tabular}{cllr}
\hline
&\textbf{Analysis} & \textbf{Signal} & \textbf{Avg. $\Delta$ F1} \\
\hline
\multirow[c]{4}{*}{\rotatebox{90}{\textit{Ablation}}}
& LingRF & Lexical diversity & +0.038 \\
& LingRF & Functional markers & +0.004 \\
& L+P Old-prob & Lexical diversity & +0.016 \\
& L+P Multi & Lexical diversity & +0.012 \\
\hline
\multirow[c]{3}{*}{\rotatebox{90}{\textit{Permut.}}}
& L+P Multi & PredOut probabilities & +0.299 \\
& L+P Multi & Lexical diversity & +0.035 \\
& L+P Multi & Original ling. features & +0.016 \\
\hline
\end{tabular}
}
\caption{Diagnostic summary for RQ2 on the test set (average macro $F_1$ drops). L+P denotes LingRF+PredOut, Old-prob denotes the original probabilistic models, and Multi denotes the multilingual configuration.}
\label{tab:rq2_stylometric_diagnostics}
\end{table}
Among the added stylometric groups, lexical diversity is the strongest signal. In LingRF, removing it causes the largest performance drop; after adding PredOut probabilities its contribution shrinks but remains the largest among the added groups. Lexical diversity measures may capture repetition and vocabulary patterns common in MGTs that are not fully reflected by probabilistic features alone. Permutation importance confirms the overall ordering: the final multilingual model is dominated by PredOut probabilities ($+0.299$), an order of magnitude above any stylometric group, with lexical diversity ($+0.035$) the strongest non-probabilistic contributor and slightly above the original linguistic features. This is consistent with the small feature-set effect seen in Table~\ref{tab:lingrf_predout_rq2}: the stylometric extension helps mainly through lexical diversity, but its contribution is minor next to the predictability signal.

\textbf{SHAP analysis}. The grouped SHAP plots in Appendix~\ref{sec:app_shap} (Figures~\ref{fig:shap_lingrf_grouped} and~\ref{fig:shap_predout_grouped}) show the same pattern: lexical diversity is the strongest added stylometric group in LingRF, whereas PredOut probabilities dominate the final multilingual model. Signed SHAP values further show that lexical diversity features are class-dependent rather than universal markers. For example, higher \texttt{root\_ttr} supports the AI class in Subtask~1, but separates generator groups differently in Subtask~2. Overall, the added stylometric features provide complementary information mainly through lexical diversity, but the final LingRF + PredOut multilingual model remains primarily driven by probabilistic outputs, and the marginal effect of the stylometric extension falls within seed variance once these outputs are included.

\section{Discussion and Conclusion}

We replicated and extended a hybrid system from the AuTexTification shared task, focusing on three aspects: (i) reproduction of the original results, (ii) the effect of different base language models, and (iii) the impact of additional stylistic features. The experiments showed practical issues with reproducibility and confirmed that feature design strongly affects performance.

\textbf{Reproducibility of experiments.} Even though we followed the description in the paper and used the available repository, we did not obtain exactly the same results. The differences were caused by factors such as different data splits, missing language models, early stopping settings, and changes in external tools. This shows that small implementation details can influence final scores. To support reproducibility, future work should release the exact code version together with clearly documented training, preprocessing, and model selection.

\textbf{Choice of models and feature design.} Replacing older models with newer multilingual ones allowed us to build one shared configuration for both languages and subtasks. Averaged over three seeds, this shared configuration performed comparably to the language-specific baseline on binary detection and improved most clearly on model attribution (Subtask~2). The results suggest that multilingual models can be a practical alternative to language-specific solutions, without claiming a uniform improvement across tasks. This is further supported by our evaluation outside the AuTexTification benchmark (see Appendix~\ref{sec:multitude_appendix} for evaluation on MULTITuDE\_v3 by~\citealp{la-cava-etal-2026-authorship}).

\textbf{Stylometric features provide a small, complementary signal.} The added stylometric features raised the mean score of the standalone linguistic model, but this gain was small relative to seed variance and could not be clearly separated from noise across three runs. Lexical diversity had the strongest effect in both the ablation and SHAP analyses. In the combined model, PredOut probabilities remained the dominant signal, especially in Subtask~1, where predictability-based features already distinguish well between human-written and generated texts. In Subtask~2, the stylometric features contributed relatively more, suggesting that attribution requires finer stylistic discrimination between generators. Overall, the results show that performance depends not only on model architecture, but also on implementation details, model selection, and feature design.

\section*{Limitations}

\textbf{Reproducibility and replacement of unavailable models.} In replication baseline, we focused on reproducing the original system. However, some models used in the baseline were no longer available and had to be replaced with suitable alternatives. Other researchers might choose different replacement models or configurations, which could lead to different results.

\textbf{Limited selection of base language models.} RQ1 focused on the selection of multilingual encoders and generative model groups. We tested only the selected subset of models, which does not cover the full range of available architectures or the most recent models. Other models or configurations could lead to different outcomes.

\textbf{Model capacity versus model family.} The multilingual probabilistic groups differ from the original models in both family and parameter count. Our design varies these factors together, so the observed gains reflect the group composition as a whole rather than an isolated effect of model capacity. A controlled same-family comparison across sizes is left for future work.

\textbf{Limited set of linguistic features.} In RQ2, we extended the original set of stylometric features, which the original paper described as its main limitation. However, our feature set is not exhaustive. Although, it provides a deeper analysis than the original set, there are still additional lexical, syntactic, or discourse-level features that could further improve performance. 

\textbf{Limited data.} Our main experiments focus on the AuTexTification 2023 shared-task dataset. Although we include a complementary MULTITuDE\_v3 mAA evaluation in Appendix~\ref{sec:multitude_appendix}, this does not constitute a full cross-benchmark validation. The results may depend on the domains, languages, and generators included in each benchmark. In particular, the generators in AuTexTification reflect the period in which the shared task was created and do not include recent LLMs. Therefore, the findings may not fully generalize to newer generation models. The MULTITuDE\_v3 results also vary across languages (Appendix~\ref{sec:multitude_appendix}), so per-language performance should be read individually rather than from the average alone.

\textbf{Comparison with shared task systems.} Our comparison with AuTexTification 2023 shared task systems should be interpreted as a post-hoc analysis rather than a blind shared-task submission. Since our experiments were conducted after the shared task, we had access to the gold labels of the test set. This allowed us to perform additional diagnostic analyses, such as SHAP-based interpretation and feature-group ablations, which would not have been possible under a blind evaluation setting without gold labels.

\section*{Ethical Considerations}

This work focuses on the analysis of text generation and detection models and does not involve personal or sensitive data. We did not use datasets containing identifiable information about individuals, and data processing complies with applicable legislation. Generative models were not used to write this paper or to design the methodology. 

\section*{Acknowledgments}
This work was partially supported by the EU NextGenerationEU through the Recovery and Resilience Plan for Slovakia under the project No. 09I01-03-V04-00059, partially by CEDMO 2.0, a project funded by the European Union under the {GA No. 101158609}, and partially by \textit{LorAI -- Low Resource Artificial Intelligence}, a project funded by Horizon Europe under \href{https://doi.org/10.3030/101136646}{GA No.101136646}. This work was also supported by the Ministry of Education, Youth and Sports of the Czech Republic through the e-INFRA CZ (ID:90254).

\bibliography{custom, anthology}

@misc{openai2023gpt4,
  title={GPT-4 Technical Report},
  author={OpenAI},
  year={2023},
  eprint={2303.08774},
  archivePrefix={arXiv},
  primaryClass={cs.CL}
}

@article{grattafiori2024llama,
  title={The llama 3 herd of models},
  author={Grattafiori, Aaron and Dubey, Abhimanyu and Jauhri, Abhinav and Pandey, Abhinav and Kadian, Abhishek and Al-Dahle, Ahmad and Letman, Aiesha and Mathur, Akhil and Schelten, Alan and Vaughan, Alex and others},
  journal={arXiv preprint arXiv:2407.21783},
  year={2024}
}

@article{zellers2019defending,
  title={Defending against neural fake news},
  author={Zellers, Rowan and Holtzman, Ari and Rashkin, Hannah and Bisk, Yonatan and Farhadi, Ali and Roesner, Franziska and Choi, Yejin},
  journal={Advances in neural information processing systems},
  volume={32},
  year={2019}
}

@article{gehrmann2019gltr,
  title={Gltr: Statistical detection and visualization of generated text},
  author={Gehrmann, Sebastian and Strobelt, Hendrik and Rush, Alexander M},
  journal={arXiv preprint arXiv:1906.04043},
  year={2019}
}

@inproceedings{mitchell2023detectgpt,
  title={Detectgpt: Zero-shot machine-generated text detection using probability curvature},
  author={Mitchell, Eric and Lee, Yoonho and Khazatsky, Alexander and Manning, Christopher D and Finn, Chelsea},
  booktitle={International conference on machine learning},
  pages={24950--24962},
  year={2023},
  organization={PMLR}
}

@article{bao2023fast,
  title={Fast-detectgpt: Efficient zero-shot detection of machine-generated text via conditional probability curvature},
  author={Bao, Guangsheng and Zhao, Yanbin and Teng, Zhiyang and Yang, Linyi and Zhang, Yue},
  journal={arXiv preprint arXiv:2310.05130},
  year={2023}
}

@inproceedings{mikros2023ai,
  title={AI-Writing Detection Using an Ensemble of Transformers and Stylometric Features.},
  author={Mikros, George K and Koursaris, Athanasios and Bilianos, Dimitrios and Markopoulos, George},
  booktitle={IberLEF@ SEPLN},
  year={2023}
}

@inproceedings{alecakir2024groningen,
  title={Groningen team a at semeval-2024 task 8: Human/machine authorship attribution using a combination of probabilistic and linguistic features},
  author={Alecakir, Huseyin and Chakraborty, Puja and Henningsson, Pontus and Van Hofslot, Matthijs and Scheuer, Alon},
  booktitle={Proceedings of the 18th International Workshop on Semantic Evaluation (SemEval-2024)},
  pages={1926--1932},
  year={2024}
}

@article{wang2024m4gt,
  title={M4gt-bench: Evaluation benchmark for black-box machine-generated text detection},
  author={Wang, Yuxia and Mansurov, Jonibek and Ivanov, Petar and Su, Jinyan and Shelmanov, Artem and Tsvigun, Akim and Afzal, Osama Mohanned and Mahmoud, Tarek and Puccetti, Giovanni and Arnold, Thomas and others},
  journal={arXiv preprint arXiv:2402.11175},
  year={2024}
}

@article{soto2024few,
  title={Few-shot detection of machine-generated text using style representations},
  author={Soto, Rafael Rivera and Koch, Kailin and Khan, Aleem and Chen, Barry and Bishop, Marcus and Andrews, Nicholas},
  journal={arXiv preprint arXiv:2401.06712},
  year={2024}
}

@misc{hans2024spottingllmsbinocularszeroshot,
      title={Spotting LLMs With Binoculars: Zero-Shot Detection of Machine-Generated Text}, 
      author={Abhimanyu Hans and Avi Schwarzschild and Valeriia Cherepanova and Hamid Kazemi and Aniruddha Saha and Micah Goldblum and Jonas Geiping and Tom Goldstein},
      year={2024},
      eprint={2401.12070},
      archivePrefix={arXiv},
      primaryClass={cs.CL},
      url={https://arxiv.org/abs/2401.12070}, 
}

@article{su2023detectllm,
  title={Detectllm: Leveraging log rank information for zero-shot detection of machine-generated text},
  author={Su, Jinyan and Zhuo, Terry Yue and Wang, Di and Nakov, Preslav},
  journal={arXiv preprint arXiv:2306.05540},
  year={2023}
}

@inproceedings{przybyla2023ve,
  title={I've Seen Things You Machines Wouldn't Believe: Measuring Content Predictability to Identify Automatically-Generated Text.},
  author={Przybyla, Piotr and Duran-Silva, Nicolau and G{\'o}mez, Santiago Egea},
  booktitle={IberLEF@ SEPLN},
  year={2023}
}

@misc{schaaff2023classificationhumanaigeneratedtexts,
      title={Classification of Human- and AI-Generated Texts for English, French, German, and Spanish}, 
      author={Kristina Schaaff and Tim Schlippe and Lorenz Mindner},
      year={2023},
      eprint={2312.04882},
      archivePrefix={arXiv},
      primaryClass={cs.CL},
      url={https://arxiv.org/abs/2312.04882}, 
}

@article{guo2024detective,
  title={Detective: Detecting ai-generated text via multi-level contrastive learning},
  author={Guo, Xun and He, Yongxin and Zhang, Shan and Zhang, Ting and Feng, Wanquan and Huang, Haibin and Ma, Chongyang},
  journal={Advances in Neural Information Processing Systems},
  volume={37},
  pages={88320--88347},
  year={2024}
}

@inproceedings{zhang2024nuanced,
  title={Nuanced Multi-class Detection of Machine-Generated Scientific Text},
  author={Zhang, Shiyuan and Ge, Yubin and Liu, Xiaofeng},
  booktitle={Proceedings of the 38th Pacific Asia Conference on Language, Information and Computation},
  pages={119--130},
  year={2024}
}

@article{wang2024trace,
  title={TRACE: TRansformer-based Attribution using Contrastive Embeddings in LLMs},
  author={Wang, Cheng and Lu, Xinyang and Ng, See-Kiong and Low, Bryan Kian Hsiang},
  journal={arXiv preprint arXiv:2407.04981},
  year={2024}
}

@article{kuznetsov2024robust,
  title={Robust ai-generated text detection by restricted embeddings},
  author={Kuznetsov, Kristian and Tulchinskii, Eduard and Kushnareva, Laida and Magai, German and Barannikov, Serguei and Nikolenko, Sergey and Piontkovskaya, Irina},
  journal={arXiv preprint arXiv:2410.08113},
  year={2024}
}

@article{kadhim2025adversarial,
  title={Adversarial attacks on AI-generated text detection models: A token probability-based approach using embeddings},
  author={Kadhim, Ahmed K and Jiao, Lei and Shafik, Rishad and Granmo, Ole-Christoffer},
  journal={arXiv preprint arXiv:2501.18998},
  year={2025}
}

@article{conneau2019unsupervised,
  title={Unsupervised cross-lingual representation learning at scale},
  author={Conneau, Alexis and Khandelwal, Kartikay and Goyal, Naman and Chaudhary, Vishrav and Wenzek, Guillaume and Guzm{\'a}n, Francisco and Grave, Edouard and Ott, Myle and Zettlemoyer, Luke and Stoyanov, Veselin},
  journal={arXiv preprint arXiv:1911.02116},
  year={2019}
}

@article{shliazhko2024mgpt,
  title={mgpt: Few-shot learners go multilingual},
  author={Shliazhko, Oleh and Fenogenova, Alena and Tikhonova, Maria and Kozlova, Anastasia and Mikhailov, Vladislav and Shavrina, Tatiana},
  journal={Transactions of the Association for Computational Linguistics},
  volume={12},
  pages={58--79},
  year={2024},
  publisher={MIT Press One Broadway, 12th Floor, Cambridge, Massachusetts 02142, USA~…}
}

@inproceedings{chiruzzo2024overview,
  title={Overview of IberLEF 2024: natural language processing challenges for Spanish and other Iberian languages},
  author={Chiruzzo, Luis and Jim{\'e}nez-Zafra, Salud Mar{\'\i}a and Rangel, Francisco},
  booktitle={Proceedings of the Iberian Languages Evaluation Forum (IberLEF 2024), co-located with the 40th Conference of the Spanish Society for Natural Language Processing (SEPLN 2024), CEUR-WS. org},
  year={2024}
}

@article{sarvazyan2023overview,
  title={Overview of {AuTexTification} at {IberLEF} 2023: Detection and Attribution of Machine-Generated Text in Multiple Domains},
  author={Sarvazyan, Areg Mikael and Gonz{\'a}lez, Jos{\'e} {\'A}ngel and Franco-Salvador, Marc and Rangel, Francisco and Chulvi, Berta and Rosso, Paolo},
  journal={Procesamiento del Lenguaje Natural},
  volume={71},
  pages={275--288},
  year={2023}
}

@inproceedings{la-cava-etal-2026-authorship,
    title = "Authorship Attribution in Multilingual Machine-Generated Texts",
    author = "La Cava, Lucio  and
      Macko, Dominik  and
      Moro, Robert  and
      Srba, Ivan  and
      Tagarelli, Andrea",
    editor = "Liakata, Maria  and
      Moreira, Viviane P.  and
      Zhang, Jiajun  and
      Jurgens, David",
    booktitle = "Proceedings of the 64th Annual Meeting of the {A}ssociation for {C}omputational {L}inguistics (Volume 1: Long Papers)",
    month = jul,
    year = "2026",
    address = "San Diego, California, United States",
    publisher = "Association for Computational Linguistics",
    url = "https://aclanthology.org/2026.acl-long.2091/",
    doi = "10.18653/v1/2026.acl-long.2091",
    pages = "45136--45152",
    ISBN = "979-8-89176-390-6",
}

@inproceedings{Shamardina_2022,
   title={Findings of the The {RuATD} Shared Task 2022 on Artificial Text Detection in {Russian}},
   url={http://dx.doi.org/10.28995/2075-7182-2022-21-497-511},
   DOI={10.28995/2075-7182-2022-21-497-511},
   booktitle={Computational Linguistics and Intellectual Technologies},
   publisher={RSUH},
   author={Shamardina, Tatiana and Mikhailov, Vladislav and Chernianskii, Daniil and Fenogenova, Alena and Saidov, Marat and Valeeva, Anastasiya and Shavrina, Tatiana and Smurov, Ivan and Tutubalina, Elena and Artemova, Ekaterina},
   year={2022},
   month=jun }

@article{luo2024understanding,
  title={From understanding to utilization: A survey on explainability for large language models},
  author={Luo, Haoyan and Specia, Lucia},
  journal={arXiv preprint arXiv:2401.12874},
  year={2024}
}

@article{lundberg2017unified,
  title={A unified approach to interpreting model predictions},
  author={Lundberg, Scott M and Lee, Su-In},
  journal={Advances in neural information processing systems},
  volume={30},
  year={2017}
}

@article{macko2025mdok,
  title={mdok of KInIT: robustly fine-tuned LLM for binary and multiclass AI-generated text detection},
  author={Macko, Dominik},
  journal={arXiv preprint arXiv:2506.01702},
  year={2025}
}

@article{he2021debertav3,
  title={Debertav3: Improving deberta using electra-style pre-training with gradient-disentangled embedding sharing},
  author={He, Pengcheng and Gao, Jianfeng and Chen, Weizhu},
  journal={arXiv preprint arXiv:2111.09543},
  year={2021}
}

@inproceedings{
melis2018on,
title={On the State of the Art of Evaluation in Neural Language Models},
author={Gábor Melis and Chris Dyer and Phil Blunsom},
booktitle={International Conference on Learning Representations},
year={2018},
url={https://openreview.net/forum?id=ByJHuTgA-},
}

@inproceedings{sharma2024team,
  title={Team innovative at semeval-2024 task 8: Multigenerator, multidomain, and multilingual black-box machine-generated text detection},
  author={Sharma, Surbhi and Mansuri, Irfan},
  booktitle={Proceedings of the 18th International Workshop on Semantic Evaluation (SemEval-2024)},
  pages={1172--1176},
  year={2024}
}

@inproceedings{petukhova2024petkaz,
  title={PetKaz at SemEval-2024 Task 8: Can Linguistics Capture the Specifics of LLM-generated Text?},
  author={Petukhova, Kseniia and Kazakov, Roman and Kochmar, Ekaterina},
  booktitle={Proceedings of the 18th International Workshop on Semantic Evaluation (SemEval-2024)},
  pages={1140--1147},
  year={2024}
}

@inproceedings{darwinkel2024groningen,
  title={Groningen Group E at SemEval-2024 Task 8: Detecting machine-generated texts through pre-trained language models augmented with explicit linguistic-stylistic features},
  author={Darwinkel, Patrick and Van Vaals, Sijbren and Van Der Holt, Marieke and Van Houten, Jarno},
  booktitle={Proceedings of the 18th International Workshop on Semantic Evaluation (SemEval-2024)},
  pages={1006--1014},
  year={2024}
}

\appendix

\section{Original Study To Be Replicated}
\label{sec:appendix}

The original study by \citet{przybyla2023ve} focuses on detecting automatically generated texts using probabilistic and textual features. The authors hypothesized that automatically generated texts differ in their token probability distributions from human-written texts, and that these differences can be used for classification. Their system was submitted to the AuTexTification 2023 shared task, which comprised two subtasks. Subtask 1 addressed binary classification, distinguishing between human-written and machine-generated texts in English and Spanish. Subtask 2 extended this setup to a multi-way classification problem, where the objective was to attribute each machine-generated text to one of several specific generative models. 

\subsection{System Overview}

\begin{figure*}[t]
\centering
\includegraphics[width=0.99\linewidth]{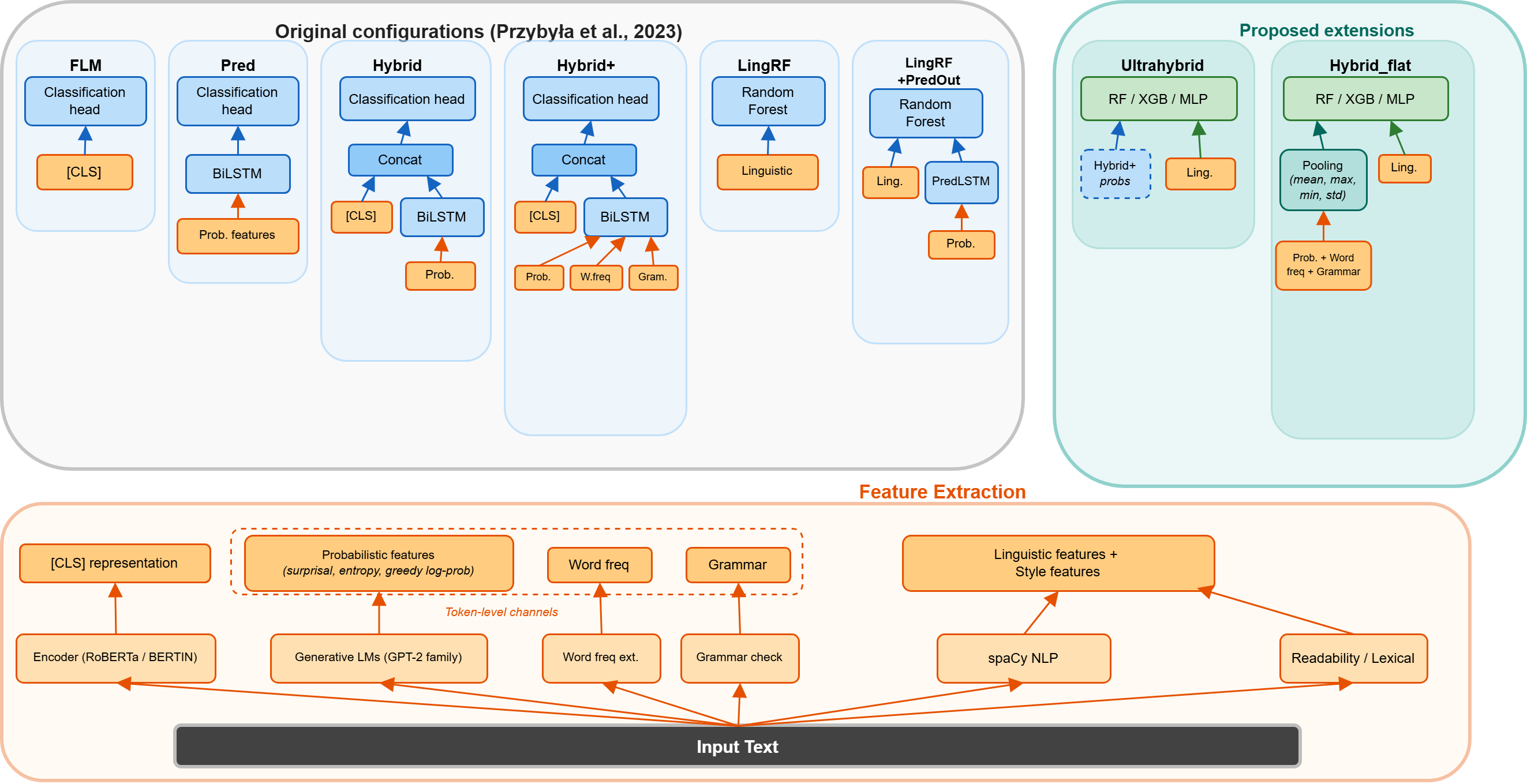}
\caption{Overview of the architecture proposed by \citet{przybyla2023ve}. The upper part shows the evaluated model configurations, while the lower part illustrates the feature extraction pipeline used to derive token-level probabilistic features and document-level linguistic features.}
\label{fig:system_overview}
\end{figure*}

The authors proposed a hybrid system combining token-level predictability features derived from generative language models with word-level linguistic features and contextual representations from a fine-tuned transformer model. The central assumption is that machine-generated texts tend to be less surprising and more predictable for language models, while human-written texts exhibit higher entropy and variability. The overall architecture and feature extraction pipeline are illustrated in Figure~\ref{fig:system_overview}.

The system is composed of the following feature groups (see Figure~\ref{fig:system_overview}):

\begin{itemize}
    \item \textbf{Predictability-based features}: for each token position, the log-probability of the observed token, the log-probability of the most likely token, and the entropy of the token probability distribution, computed using generative language models (GPT-2 variants).
    
    \item \textbf{Word-level features}: logarithmic word frequency derived from the Google Books Ngrams corpus and grammatical correctness indicators obtained using LanguageTool.
    
    \item \textbf{Text-level linguistic features}: aggregated statistics of part-of-speech tags, morphological categories, dependency relations, named entities, and rare-word indicators.
    
    \item \textbf{Contextual representations}: pooled [CLS] representations extracted from fine-tuned RoBERTa models.
\end{itemize}

\subsection{Methodology}

As mentioned above, the authors participated in the AuTexTification 2023 shared task and used the dataset provided within the competition. A more detailed description of the dataset is presented in Section~\ref{chap:meth} of our work.

\textbf{Text processing.} Each text was processed using GPT-2 language models of different sizes to compute token-level predictability features. For each token position (up to length 128), the following quantities were extracted: (i) the log-probability of the observed token, (ii) the log-probability of the most likely token, and (iii) the entropy of the token probability distribution. Instead of aggregating these values into a single perplexity score, the full sequence of token-level features was retained.

In addition, two types of word-level features were introduced:

\begin{itemize}
    \item \textbf{Word frequency} -- obtained from the Google Books Ngrams v3 dataset for English and Spanish, aligned with the language model tokenization;
    \item \textbf{Grammatical correctness} -- obtained using LanguageTool, where tokens were marked according to whether they were preserved or modified by the grammar checker.
\end{itemize}

To enrich the text representation, the authors fine-tuned RoBERTa (RoBERTa-base for English and RoBERTa-base-BNE for Spanish) and used the pooled [CLS] vector as a contextual embedding. They also extracted linguistic information at the document level, including part-of-speech tags, dependency relations, morphological categories, and named entities. These annotations were converted into aggregated statistics.

\textbf{Model configuration.} The main neural component is a bidirectional LSTM (BiLSTM) operating on sequences of token-level predictability features obtained from several GPT-2 models (DistilGPT-2, GPT-2, GPT-2 Medium, and GPT-2 Large). The hidden size is 64 in each direction, yielding a 128-dimensional representation after concatenation. This vector can be combined with the RoBERTa embedding and passed to a classification layer. In addition, a Random Forest classifier is trained on the aggregated linguistic features.

\textbf{Model variants.} Based on their overall configuration, the authors defined the following variants for evaluation:
\begin{itemize}
    \item \textbf{Pred}: a BiLSTM network using predictability features,
    \item \textbf{FLM}: only the fine-tuned language model, used as a baseline,
    \item \textbf{Pred + FLM}: a linear classifier combining representations from Pred and FLM, referred to as \textit{Hybrid},
    \item \textbf{Pred + FLM + Add}: as above, but including additional token-level features, referred to as \textit{Hybrid+},
    \item \textbf{LingRF}: a Random Forest trained on linguistic features,
    \item \textbf{LingRF + PredOut}: a Random Forest using linguistic features and probabilities returned by the BiLSTM network operating on the predictability features, referred to as \textit{LinguisticRF}.
\end{itemize}

\textbf{Implementation.} The neural models were implemented in PyTorch, and pretrained language models were loaded via HuggingFace Transformers. The Adam optimizer was used with a learning rate of $10^{-3}$ for the BiLSTM and $2 \times 10^{-5}$ for fine-tuning RoBERTa. In the combined setting, RoBERTa was frozen for the first five epochs and then trained jointly with the rest of the network using the smaller learning rate.

Topic-based experiments were trained for 10 epochs. For the final submission, the data were split into 80\% training and 20\% development sets and trained for up to 20 epochs. Model selection was based on development macro F1-score, with early stopping at the first epoch reaching at least $(f_{\max} - 0.01)$.

Random Forest classifiers were implemented in Scikit-learn. After tuning, 200 trees with maximum depth 60 were used for all settings. Linguistic features were annotated using spaCy (\texttt{en\_core\_web\_sm}, \texttt{es\_core\_news\_sm}) and aggregated into document-level statistics. Word-frequency features were derived from the Google Books Ngrams corpus. The official implementation is available at \url{https://github.com/piotrmp/autext}.

Because the original paper does not provide a complete list of linguistic features, we also compare the reported feature counts with the counts obtained in our reproduction. Table~\ref{tab:feat_count} shows that the reconstructed counts closely match the original paper for English, while larger differences appear for Spanish.

\begin{table}[!t]
\centering
\setlength{\tabcolsep}{3.5pt}
\resizebox{0.8\columnwidth}{!}{%
\begin{tabular}{lllrr}
\hline
\textbf{Subtask} & \textbf{Lang} & \textbf{Variant} & \textbf{Our} & \textbf{Orig} \\
\hline
\multirow{4}{*}{S1} 
& \multirow{2}{*}{en} & LingRF          & 163 & 165 \\
&                      & LingRF+PredOut  & 165 & 165 \\
& \multirow{2}{*}{es} & LingRF          & 311 & 299 \\
&                      & LingRF+PredOut  & 313 & 299 \\
\hline
\multirow{4}{*}{S2} 
& \multirow{2}{*}{en} & LingRF          & 163 & 163 \\
&                      & LingRF+PredOut  & 169 & 163 \\
& \multirow{2}{*}{es} & LingRF          & 307 & 288 \\
&                      & LingRF+PredOut  & 313 & 288 \\
\hline
\end{tabular}%
}
\caption{Comparison of linguistic feature counts between our reproduction (Our) and the original paper (Orig) for LingRF and LingRF+PredOut.}
\label{tab:feat_count}
\end{table}

\section{Additional Stylometric Features}
\label{sec:app_features}

The additional stylometric features introduced as part of RQ2 are summarized in Table~\ref{tab:new_features}. These features extend the original setup by incorporating measures of lexical diversity, sentence structure, repetition patterns, word-level statistics, functional and stylistic markers, readability metrics, and punctuation usage, thereby providing a more fine-grained characterization of writing style.

All features are extracted at the document level using lightweight rule-based methods. Texts are tokenized with regex-based word splitting and sentence segmentation based on terminal punctuation. Lexical diversity, repetition, and word-level statistics are computed from token frequency distributions, while stylistic marker ratios are obtained by matching tokens against curated word lists. Readability metrics are calculated using the \texttt{textstat} library.

The curated word lists (transition, hedge, function, formal words, and first-person pronouns) are language-specific and were compiled manually for this work, following common stopword and discourse-marker conventions and adapted for each language. Their sizes are reported in Table~\ref{tab:app_wordlists}; the complete lists are available in the released code.

\begin{table}[!t]
\centering
\small
\begin{tabular}{lcc}
\hline
\textbf{Word list} & \textbf{English} & \textbf{Spanish} \\
\hline
Transition words     & 31  & 26 \\
Hedge words          & 40  & 24 \\
Function words       & 118 & 94 \\
Formal words         & 58  & 44 \\
First-person pronouns & 10 & 10 \\
\hline
\end{tabular}
\caption{Sizes of the curated word lists used by the functional and stylistic
marker features.}
\label{tab:app_wordlists}
\end{table}

\begin{table*}[!t]
\centering
\footnotesize
\setlength{\tabcolsep}{4pt}
\renewcommand{\arraystretch}{1.1}
\resizebox{0.8\linewidth}{!}{%
\begin{tabular}{p{3.2cm} p{3.2cm} p{6cm}}
\hline
\textbf{Category} & \textbf{Feature} & \textbf{Description} \\
\hline

Lexical Diversity & \texttt{ttr} & Type--token ratio (unique tokens / total tokens). \\
& \texttt{root\_ttr} & TTR normalized by the square root of token count. \\
& \texttt{log\_ttr} & Logarithmic variant of TTR. \\
& \texttt{hapax\_ratio} & Ratio of words occurring once in the document. \\
& \texttt{dis\_legomena\_ratio} & Ratio of words occurring exactly twice. \\
& \texttt{rare\_word\_burstiness} & Burstiness (clustering) of rare words across the text. \\

Sentence Structure & \texttt{avg\_sentence\_length} & Average tokens per sentence. \\
& \texttt{sentence\_length\_std} & Standard deviation of sentence length. \\
& \texttt{sentence\_length\_cv} & Coefficient of variation of sentence length. \\
& \texttt{sentence\_count} & Number of sentences in the document. \\

Repetition Patterns & \texttt{bigram\_repetition} & Ratio of excess repeated bigrams to total bigrams. \\
& \texttt{trigram\_repetition} & Ratio of excess repeated trigrams to total trigrams. \\

Word-Level Statistics & \texttt{avg\_word\_length} & Average token length in characters. \\
& \texttt{word\_length\_std} & Standard deviation of token length. \\
& \texttt{word\_count} & Total number of tokens in the document. \\

Functional \& Stylistic Markers & \texttt{function\_word\_ratio} & Ratio of function words to all tokens. \\
& \texttt{transition\_word\_ratio} & Ratio of discourse transition words. \\
& \texttt{hedge\_word\_ratio} & Ratio of hedging expressions (e.g., \emph{maybe}, \emph{possibly}). \\
& \texttt{first\_person\_ratio} & Ratio of first-person pronouns. \\
& \texttt{formal\_word\_ratio} & Ratio of formal register vocabulary. \\

Readability Metrics & \texttt{flesch\_reading\_ease} & Flesch Reading Ease score. \\
& \texttt{flesch\_kincaid\_grade} & Flesch--Kincaid grade level. \\

Punctuation Usage & \texttt{punctuation\_ratio} & Ratio of punctuation characters to total characters. \\
& \texttt{comma\_ratio} & Average number of commas per sentence. \\
& \texttt{exclamation\_ratio} & Average number of exclamation marks per sentence. \\
& \texttt{question\_ratio} & Average number of question marks per sentence. \\

\hline
\end{tabular}
}
\caption{Additional stylometric features introduced in this study.}
\label{tab:new_features}
\end{table*}

\section{Additional results}

In this section, we present additional experiments and comparisons that were not included in the main analysis due to space limitations. These results are not directly tied to the main research questions, but provide additional context for robustness, alternative configurations, comparison with external systems, and cross-dataset behavior.

We evaluate two additional configurations derived from the baseline, namely \textit{Ultrahybrid} and \textit{Hybrid\_flat}. Their architectures and feature processing are described in the following subsection.

We also provide a more detailed comparison between the original baseline and the proposed multilingual configuration, focusing on model robustness across different random seeds and the computational cost of training and inference.

\subsection{New Configurations} \label{chap:new_models}

As part of the original extension, we also explored additional model configurations. This section describes their structure and the way input features are processed.

The \textbf{Ultrahybrid} configuration is implemented as a two-stage model. In the first stage, the Hybrid+ model is trained end-to-end. Sequential token-level features (probabilistic channels, token frequency, and grammar flag) are processed by the BiLSTM component, while the input text is processed by the encoder component. After training, the softmax class probabilities are extracted for each document.
In the second stage, these probabilities are concatenated with document-level linguistic features. The resulting feature matrix is used to train traditional machine-learning classifiers: Random Forest, XGBoost, and a multilayer perceptron.

The results of the Ultrahybrid configuration are reported in Table \ref{tab:hybrid_ultrahybrid_test_all}.
\begin{table}[!t]
\centering
\small
\resizebox{\columnwidth}{!}{%
\begin{tabular}{llcccc}
\hline
 &  & \textbf{Hybrid+} & \multicolumn{3}{c}{\textbf{Ultrahybrid}} \\
\cmidrule(lr){4-6}
\textbf{Subtask} & \textbf{Lang}
 &  & \textbf{RF} & \textbf{XGB} & \textbf{MLP} \\
\hline
\multicolumn{6}{c}{\textit{Baseline prob models}} \\
S1 & en & \textbf{0.796} & \textbf{0.796} & 0.795 & 0.779 \\
S1 & es & \textbf{0.743} & 0.720 & 0.726 & 0.713 \\
S2 & en & 0.586 & 0.599 & 0.601 & 0.591 \\
S2 & es & 0.567 & 0.591 & 0.590 & 0.598 \\
\hline
\multicolumn{6}{c}{\textit{Multilingual prob models}} \\
S1 & en & 0.765 & 0.771 & 0.777 & 0.780 \\
S1 & es & 0.704 & 0.731 & 0.738 & 0.717 \\
S2 & en & 0.595 & 0.636 & 0.635 & \textbf{0.637} \\
S2 & es & 0.635 & 0.647 & \textbf{0.652} & 0.645 \\
\hline
\end{tabular}%
}
\caption{Test macro $F_1$ scores for Ultrahybrid (baseline vs.\ multilingual).}
\label{tab:hybrid_ultrahybrid_test_all}
\end{table}
From the table, we can see that in the baseline setting, UltraHybrid does not improve the results for Subtask 1 in either language. The class probabilities produced by Hybrid+ already represent a very strong signal, and the second-stage classifiers cannot significantly enhance it. The additional document-level features therefore do not provide useful complementary information. For the more challenging Subtask 2, however, UltraHybrid consistently outperforms the original Hybrid+. In the multilingual setting, the benefit of UltraHybrid is more visible. In all four combinations (Subtask 1/2 × EN/ES), it achieves higher scores than Hybrid+. This suggests that in this setting, the first-stage probabilities are not as dominant, and the second-stage classifier can make better use of its additional features.

The \textbf{Hybrid\_flat} configuration uses the same token-level features as Hybrid+, but without the sequential component and without the FLM component. The same probabilistic channels are used as in previous experiments, together with the logarithmic token frequency and a binary grammatical correctness flag.

Instead of processing the full sequence with a BiLSTM, each real token-level channel is aggregated to the document level using four statistics: mean, maximum, minimum, and standard deviation. Let $x_{i}^{(k)}$ denote the value of the $k$-th channel for the $i$-th valid token in a document, and let $N$ be the number of valid tokens given by the mask. For each channel $k$, the following aggregated features are computed:

\[
\mu_k = \frac{1}{N} \sum_{i=1}^{N} x_{i}^{(k)},
\]

\[
\max_k = \max_{1 \le i \le N} x_{i}^{(k)},
\]

\[
\min_k = \min_{1 \le i \le N} x_{i}^{(k)},
\]

\[
\sigma_k = \sqrt{\frac{1}{N} \sum_{i=1}^{N} \left(x_{i}^{(k)} - \mu_k\right)^2}.
\]

If the number of real token-level channels is denoted by $K$, the dimensionality of the aggregated part is

\[
4 \times K.
\]

In the English baseline setting (12 probabilistic channels, 1 frequency, and 1 grammar channel), we have $K = 14$, which results in $4 \times 14 = 56$ aggregated features.

These aggregated features are concatenated with standard linguistic document-level features, forming a fixed-length vector representing the whole document. This vector is used as input to the final classifier. As in the Ultrahybrid configuration, we experimented with Random Forest, XGBoost, and a multilayer perceptron. The results are shown in the table \ref{tab:hybridflat}.

\begin{table}[!t]
\centering
\resizebox{0.7\columnwidth}{!}{%
\begin{tabular}{lcccc}
\hline
\multicolumn{5}{c}{\textit{Baseline prob models}} \\
\hline
\textbf{Model} & \textbf{S1-en} & \textbf{S1-es} & \textbf{S2-en} & \textbf{S2-es} \\
\hline
RF  & 0.825 & 0.781 & 0.502 & 0.502 \\
XGB & 0.816 & 0.789 & 0.536 & 0.532 \\
MLP & 0.813 & 0.749 & 0.547 & 0.521 \\
\hline
\multicolumn{5}{c}{\textit{Multilingual prob models}} \\
\hline
\textbf{Model} & \textbf{S1-en} & \textbf{S1-es} & \textbf{S2-en} & \textbf{S2-es} \\
\hline
RF  & 0.843 & 0.875 & 0.533 & 0.549 \\
XGB & \textbf{0.847} & \textbf{0.877} & 0.566 & \textbf{0.585} \\
MLP & 0.808 & 0.782 & \textbf{0.595} & 0.568 \\
\hline
\end{tabular}
}
\caption{Test macro $F_1$ scores across subtasks and languages for Hybrid\_flat.}
\label{tab:hybridflat}
\end{table}

As shown in the table, we compared the proposed configuration in two settings: baseline (probabilistic models used in replication baseline to replicate the original study) and multilingual (probabilistic models proposed as an improved configuration in RQ1). The results show that the configuration using multilingual probabilistic models achieves better results than the baseline, especially for tasks focused on Spanish. Regarding the individual models, XGBoost achieves the best results in three out of four cases, while MLP performs best in one case. The proposed configuration achieves high results overall and, particularly for Subtask 1, clearly outperforms the original solution that we replicated (see Table \ref{tab:rq1_full}).

\subsection{Robustness and Time Complexity} \label{timecomplexity_baseline}

In this section, we analyze the stability of the proposed multilingual configuration (mDeBERTa-v3 + Large group) from RQ1. The goal is to verify whether the observed improvements are not caused by random initialization and whether the results are consistent across different seeds.

We repeated the experiments with three random seeds (10, 11, 12) and report the mean performance and standard deviation for both the baseline and the multilingual configuration. In addition to robustness, we also analyze time complexity by comparing the total training and inference time of the baseline and the proposed multilingual solution. The results are shown in Table~\ref{tab:rq2_robustness_time}.
\begin{table}[!t]
\centering
\small
\setlength{\tabcolsep}{4pt}
\resizebox{0.8\columnwidth}{!}{%
\begin{tabular}{lllc}
\hline
\textbf{Lang} & \textbf{Model} & \textbf{Test macro $F_1$} & \textbf{Time (min)} \\
\hline
\multicolumn{4}{c}{\textit{Baseline prob models}} \\
\cmidrule(lr){1-4}
\multicolumn{4}{l}{\textbf{S1}} \\
en & Hybrid  & 0.788 $\pm$ 0.034 & 76.1 \\
en & Hybrid+ & 0.793 $\pm$ 0.077 & 77.4 \\
es & Hybrid  & 0.721 $\pm$ 0.006 & 55.1 \\
es & Hybrid+ & 0.728 $\pm$ 0.012 & 55.4 \\
\cmidrule(lr){1-4}
\multicolumn{4}{l}{\textbf{S2}} \\
en & Hybrid  & 0.580 $\pm$ 0.011 & 47.0 \\
en & Hybrid+ & 0.588 $\pm$ 0.002 & 47.5 \\
es & Hybrid  & 0.589 $\pm$ 0.003 & 37.5 \\
es & Hybrid+ & 0.573 $\pm$ 0.011 & 37.8 \\
\hline
\multicolumn{4}{c}{\textit{Multilingual prob models}} \\
\cmidrule(lr){1-4}
\multicolumn{4}{l}{\textbf{S1}} \\
en & Hybrid  & 0.828 $\pm$ 0.023 & 122.1 \\
en & Hybrid+ & 0.790 $\pm$ 0.031 & 123.5 \\
es & Hybrid  & 0.747 $\pm$ 0.044 & 119.5 \\
es & Hybrid+ & 0.769 $\pm$ 0.047 & 120.2 \\
\cmidrule(lr){1-4}
\multicolumn{4}{l}{\textbf{S2}} \\
en & Hybrid  & 0.632 $\pm$ 0.009 & 75.3 \\
en & Hybrid+ & 0.620 $\pm$ 0.017 & 76.1 \\
es & Hybrid  & 0.639 $\pm$ 0.008 & 75.7 \\
es & Hybrid+ & 0.637 $\pm$ 0.002 & 76.2 \\
\hline
\end{tabular}%
}
\caption{Robustness across three runs (test macro $F_1$: mean $\pm$ std). Runtime is reported for a single run (seed=10).}
\label{tab:rq2_robustness_time}
\end{table}

From Table \ref{tab:rq2_robustness_time}, the results are stable across the three seeds. The standard deviation is low in most cases, which indicates that the performance is not strongly affected by random initialization. Higher variability appears only in some configurations in Subtask 1, but this is not a systematic effect. 

Training times are significantly lower in the baseline setting. On average, the multilingual configuration is about 80\% more time-consuming in both subtasks. This is related to the use of a larger model group and more complex probabilistic models. 

In terms of the trade-off between performance and time, the multilingual configuration provides the largest benefit in Subtask 2, where Test Macro F1 improves in both languages. In Subtask 1, the situation is less clear. Hybrid improves for EN, while Hybrid+ achieves comparable or slightly lower performance. Overall, the proposed multilingual configuration is stable and improves performance mainly in the more challenging second subtask, but at the cost of higher time complexity.

\subsection{Comparison with State-of-the-Art Detector}
\label{sec:comparison_sota}
We further compared the LingRF+PredOut configuration with mdok (based on Qwen3-4B-Base) \cite{macko2025mdok}, a strong fine-tuned detector. We evaluate the \textit{Multilingual} variant of LingRF+PredOut from Section~\ref{chap:RQ3} (multilingual probabilistic models with the extended feature set), reported as the mean over three seeds. To keep the comparison as close as possible, mdok was fine-tuned on the AuTexTification training data using the same train/dev split; we report its single-run scores. The comparison is therefore a competitiveness reference rather than a matched benchmark, since our system underwent substantially more model selection and feature engineering on this dataset, whereas mdok was evaluated with a single standard fine-tuning configuration (hyperparameters given below).

As shown in Table~\ref{tab:mdok_comparison}, the two systems occupy opposite ends of the two subtasks. On binary detection (Subtask~1) our approach outperforms mdok by a large margin in both languages, with non-overlapping ranges. On generator attribution (Subtask~2) mdok is clearly stronger in both languages. The neural detector is thus the better attributor, while our stylometry-based approach is competitive on detection and additionally provides interpretable signals, allowing us to analyze which stylistic features influence the model's decisions.

\begin{table}[!t]
\centering
\small
\setlength{\tabcolsep}{4pt}
\resizebox{0.85\columnwidth}{!}{%
\begin{tabular}{lcccc}
\hline
\textbf{Model} & \textbf{S1-en} & \textbf{S1-es} & \textbf{S2-en} & \textbf{S2-es} \\
\hline
mdok & 0.763 & 0.679 & \textbf{0.626} & \textbf{0.670} \\
LingRF+PredOut & \textbf{0.889} & \textbf{0.867} & 0.557 & 0.542 \\
\hline
\end{tabular}%
}
\caption{Comparison with the fine-tuned detector mdok on the test set (macro $F_1$). LingRF+PredOut is the \textit{Multilingual} configuration averaged over seeds 10/11/12; mdok is a single fine-tuning run on the same train/dev split.}
\label{tab:mdok_comparison}
\end{table}

\textbf{mdok fine-tuning configuration.}
For a fair comparison, we fine-tuned mdok (Qwen3-4B-Base) on the official AuTexTification training data rather than using it off-the-shelf. We use QLoRA (4-bit NF4, LoRA $r=64$, $\alpha=16$, dropout $0.1$) with a class-weighted cross-entropy loss, $3$ epochs, learning rate $2\times10^{-4}$, batch size $32$, and a $512$-token limit. For each subtask and language we hold out $20\%$ of the training set for validation (stratified, seed $10$) and keep the best-validation checkpoint. mdok was run with this single configuration and one seed, so its scores serve as a competitiveness reference rather than a tuned upper bound.

\subsection{Additional MULTITuDE Evaluation}
\label{sec:multitude_appendix}

As an additional evaluation, we tested the final multilingual configurations on MULTITuDE\_v3~\citep{la-cava-etal-2026-authorship}. This benchmark focuses on multilingual authorship attribution for machine-generated texts. Compared to the AuTexTification setting used in the main experiments, MULTITuDE includes more languages and multiple generator-specific classes. We include these results only as supplementary evidence, since this dataset was not part of the original replication setup.

\begin{table*}[!t]
\centering
\setlength{\tabcolsep}{1.0mm} 
\scalebox{0.78}{
\begin{tabular}{l||c|c|c||c|c|c||c|c|c|c|c||c|c|c||c|c|c|c||c}
\textbf{Lang. family $\rightarrow$} & \multicolumn{3}{c||}{Germanic} & \multicolumn{3}{c||}{Romance} & \multicolumn{5}{c||}{Slavic-Latin} & \multicolumn{3}{c||}{Slavic-Cyrillic} & \multicolumn{4}{c||}{Others} & \\
\textbf{Method $\downarrow$} & de & en & nl & es & pt & ro & cs & hr & pl & sk & sl & bg & ru & uk & hu & el & ar & zh & all \\
\hline
mdok & {\cellcolor[HTML]{9AD587}}0.92 & {\cellcolor[HTML]{9AD587}}0.91 & {\cellcolor[HTML]{92D183}}0.95 & {\cellcolor[HTML]{9CD687}}0.91 & {\cellcolor[HTML]{97D385}}0.93 & {\cellcolor[HTML]{93D284}}0.94 & {\cellcolor[HTML]{92D183}}0.95 &  {\cellcolor[HTML]{90D083}}0.96 &  {\cellcolor[HTML]{93D284}}0.94 &  {\cellcolor[HTML]{8DCF81}}0.97 &  {\cellcolor[HTML]{90D083}}0.95 & {\cellcolor[HTML]{95D385}}0.93 & {\cellcolor[HTML]{9AD587}}0.91 &  {\cellcolor[HTML]{95D385}}0.93 & {\cellcolor[HTML]{97D385}}0.93 &  {\cellcolor[HTML]{93D284}}0.94 & {\cellcolor[HTML]{90D083}}0.96 &  {\cellcolor[HTML]{A4D98A}}0.87 & {\cellcolor[HTML]{95D385}}0.93 \\
OTBDetector & {\cellcolor[HTML]{A2D88A}}0.87 & {\cellcolor[HTML]{B8E293}}0.78 & {\cellcolor[HTML]{9AD587}}0.91 & {\cellcolor[HTML]{A7DB8C}}0.85 & {\cellcolor[HTML]{A1D889}}0.89 & {\cellcolor[HTML]{95D385}}0.93 & {\cellcolor[HTML]{95D385}}0.93 & {\cellcolor[HTML]{95D385}}0.93 & {\cellcolor[HTML]{98D486}}0.92 & {\cellcolor[HTML]{90D083}}0.96 & {\cellcolor[HTML]{93D284}}0.94 & {\cellcolor[HTML]{97D385}}0.93 & {\cellcolor[HTML]{A4D98A}}0.87 & {\cellcolor[HTML]{9CD687}}0.91 & {\cellcolor[HTML]{9CD687}}0.91 & {\cellcolor[HTML]{98D486}}0.92 & {\cellcolor[HTML]{93D284}}0.95 & {\cellcolor[HTML]{B3E091}}0.80 & {\cellcolor[HTML]{9DD688}}0.90 \\
\textbf{Hybrid} & {\cellcolor[HTML]{A7DB8C}}0.85 & {\cellcolor[HTML]{B8E293}}0.77 & {\cellcolor[HTML]{A4D98A}}0.87 & {\cellcolor[HTML]{AEDD8E}}0.82 & {\cellcolor[HTML]{A6DA8B}}0.86 & {\cellcolor[HTML]{A1D889}}0.89 & {\cellcolor[HTML]{A1D889}}0.89 & {\cellcolor[HTML]{A1D889}}0.89 & {\cellcolor[HTML]{A2D88A}}0.88 & {\cellcolor[HTML]{93D284}}0.94 & {\cellcolor[HTML]{9CD687}}0.91 & {\cellcolor[HTML]{A4D98A}}0.87 & {\cellcolor[HTML]{ABDC8D}}0.84 & {\cellcolor[HTML]{A6DA8B}}0.86 & {\cellcolor[HTML]{A2D88A}}0.88 & {\cellcolor[HTML]{9CD687}}0.91 & {\cellcolor[HTML]{9AD587}}0.92 & {\cellcolor[HTML]{B6E192}}0.79 & {\cellcolor[HTML]{A4D98A}}0.87 \\
XLM-R-large & {\cellcolor[HTML]{B2DF90}}0.81 & {\cellcolor[HTML]{CFEC9E}}0.65 & {\cellcolor[HTML]{ABDC8D}}0.84 & {\cellcolor[HTML]{BAE394}}0.76 & {\cellcolor[HTML]{B2DF90}}0.80 & {\cellcolor[HTML]{A4D98A}}0.87 & {\cellcolor[HTML]{A1D889}}0.88 & {\cellcolor[HTML]{A1D889}}0.88 & {\cellcolor[HTML]{A2D88A}}0.88 & {\cellcolor[HTML]{97D385}}0.93 & {\cellcolor[HTML]{9DD688}}0.90 & {\cellcolor[HTML]{A4D98A}}0.87 & {\cellcolor[HTML]{B6E192}}0.78 & {\cellcolor[HTML]{ABDC8D}}0.84 & {\cellcolor[HTML]{A6DA8B}}0.86 & {\cellcolor[HTML]{A2D88A}}0.88 & {\cellcolor[HTML]{9DD688}}0.90 & {\cellcolor[HTML]{C1E698}}0.72 & {\cellcolor[HTML]{ABDC8D}}0.84 \\
RoBERTa-large & {\cellcolor[HTML]{B6E192}}0.78 & {\cellcolor[HTML]{C1E698}}0.72 & {\cellcolor[HTML]{B1DF90}}0.81 & {\cellcolor[HTML]{BDE496}}0.74 & {\cellcolor[HTML]{B2DF90}}0.80 & {\cellcolor[HTML]{A9DB8C}}0.84 & {\cellcolor[HTML]{AEDD8E}}0.83 & {\cellcolor[HTML]{ACDD8E}}0.83 & {\cellcolor[HTML]{B1DF90}}0.81 & {\cellcolor[HTML]{A9DB8C}}0.85 & {\cellcolor[HTML]{ABDC8D}}0.84 & {\cellcolor[HTML]{D2EDA0}}0.63 & {\cellcolor[HTML]{D3EDA0}}0.63 & {\cellcolor[HTML]{CCEA9D}}0.67 & {\cellcolor[HTML]{BAE394}}0.76 & {\cellcolor[HTML]{DAF0A4}}0.59 & {\cellcolor[HTML]{C5E89A}}0.70 & {\cellcolor[HTML]{D9F0A3}}0.60 & {\cellcolor[HTML]{BDE496}}0.75 \\
\textbf{LingRF + PredOut} & {\cellcolor[HTML]{C5E89A}}0.70 & {\cellcolor[HTML]{C3E799}}0.71 & {\cellcolor[HTML]{C3E799}}0.71 & {\cellcolor[HTML]{CCEA9D}}0.67 & {\cellcolor[HTML]{BFE597}}0.73 & {\cellcolor[HTML]{C3E799}}0.71 & {\cellcolor[HTML]{C5E89A}}0.70 & {\cellcolor[HTML]{C5E89A}}0.70 & {\cellcolor[HTML]{C3E799}}0.71 & {\cellcolor[HTML]{BFE597}}0.73 & {\cellcolor[HTML]{C7E99B}}0.69 & {\cellcolor[HTML]{C1E698}}0.72 & {\cellcolor[HTML]{C9E99C}}0.68 & {\cellcolor[HTML]{CFECA0}}0.66 & {\cellcolor[HTML]{D0EC9F}}0.64 & {\cellcolor[HTML]{BDE496}}0.75 & {\cellcolor[HTML]{B8E293}}0.77 & {\cellcolor[HTML]{CFEC9E}}0.65 & {\cellcolor[HTML]{C5E89A}}0.70 \\
StatEnsemble & {\cellcolor[HTML]{E7F6AD}}0.49 & {\cellcolor[HTML]{F8FCC0}}0.33 & {\cellcolor[HTML]{DEF2A7}}0.55 & {\cellcolor[HTML]{ECF7B1}}0.45 & {\cellcolor[HTML]{E9F6AF}}0.47 & {\cellcolor[HTML]{E8F6AE}}0.48 & {\cellcolor[HTML]{EDF8B2}}0.43 & {\cellcolor[HTML]{EEF9B3}}0.43 & {\cellcolor[HTML]{E5F5AC}}0.50 & {\cellcolor[HTML]{EEF9B3}}0.43 & {\cellcolor[HTML]{F9FDC2}}0.31 & {\cellcolor[HTML]{E3F4AA}}0.51 & {\cellcolor[HTML]{E8F6AE}}0.48 & {\cellcolor[HTML]{E8F6AE}}0.48 & {\cellcolor[HTML]{E5F5AC}}0.50 & {\cellcolor[HTML]{F0F9B4}}0.41 & {\cellcolor[HTML]{F2FAB5}}0.40 & {\cellcolor[HTML]{F7FCBC}}0.35 & {\cellcolor[HTML]{ECF7B1}}0.45 \\
\hline
\end{tabular}
}
\caption{Per-language macro-averaged $F_1$ scores of the selected methods on MULTITuDE\_v3 test data. Darker shades of green indicate higher scores. The performance values of existing detectors are those reported in \citep{la-cava-etal-2026-authorship}. Our detectors are boldfaced.} 
\label{mAA_multilingualresults}
\end{table*}

The MULTITuDE results suggest that the final multilingual configurations are not restricted to the English--Spanish setting of AuTexTification. Hybrid reaches 0.868 macro $F_1$ and performs better on several additional languages than on English and Spanish, while still remaining below mdok and OTBDetector. This comparison should be interpreted with the model design in mind: mdok relies on a fine-tuned LLM, and OTBDetector uses contrastive learning, whereas our Hybrid configuration uses neither of these strategies. Instead, it combines multilingual encoder representations with probabilistic token-level features. This leaves a clear path for further improvement, since LLM fine-tuning or contrastive objectives could be combined with our feature design. LingRF + PredOut is weaker than the neural baselines, but clearly outperforms StatEnsemble (0.702 vs. 0.450), indicating that PredOut-enhanced stylometric features provide a more transferable signal than statistical detector scores alone.

\section{Ablation study}
\label{sec:Abl_study}

This appendix reports the full diagnostic results for RQ2. The ablation values are mean test macro F1 drops over RF seeds 10, 11, and 12. Positive values indicate that removing or permuting a feature group lowers test macro F1 compared with the corresponding full feature set. Negative values indicate that removing or permuting the group improves performance in that setting.

For each model setting $m$, feature group $g$, and RF seed $s$, we compute the ablation drop as:
\[
\Delta_{m,g,s} =
F1_{\mathrm{full}}(m,s) - F1_{\setminus g}(m,s).
\]
The tables report $\frac{1}{3}\sum_s \Delta_{m,g,s}$. This design keeps the train/development split and the pre-computed PredOut probabilities fixed, and isolates the stability of the Random Forest contribution of each feature group.

\textbf{Leave-one-group-out ablation}.
Tables~\ref{tab:lingrf_style_ablation_drops} and~\ref{tab:predout_style_ablation_drops} report leave-one-group-out ablation for the extended stylometric feature groups. Positive values indicate a test macro F1 drop after removing a feature group, while negative values indicate a small gain after removal.

\begin{table}[!t]
\centering
\small
\setlength{\tabcolsep}{3pt}
\resizebox{\columnwidth}{!}{%
\begin{tabular}{lrrrrr}
\hline
\textbf{Removed group} & \textbf{S1-en} & \textbf{S1-es} & \textbf{S2-en} & \textbf{S2-es} & \textbf{Avg.} \\
\hline
Lexical diversity & \posdrop{+0.025} & \posdrop{+0.067} & \posdrop{+0.032} & \posdrop{+0.028} & \posdrop{+0.038} \\
Sentence structure & \posdrop{+0.007} & \negdrop{-0.016} & \posdrop{+0.002} & \posdrop{+0.005} & +0.000 \\
Repetition & \posdrop{+0.001} & \negdrop{-0.001} & \posdrop{+0.002} & \posdrop{+0.002} & \posdrop{+0.001} \\
Word-level stats & \posdrop{+0.004} & \posdrop{+0.006} & \posdrop{+0.001} & \posdrop{+0.002} & \posdrop{+0.003} \\
Functional markers & \posdrop{+0.017} & \negdrop{-0.006} & \negdrop{-0.001} & \posdrop{+0.003} & \posdrop{+0.004} \\
Readability & \posdrop{+0.011} & \negdrop{-0.005} & \negdrop{-0.001} & \posdrop{+0.008} & \posdrop{+0.003} \\
Punctuation & \posdrop{+0.001} & \negdrop{-0.010} & \negdrop{-0.002} & \posdrop{+0.006} & \negdrop{-0.001} \\
\hline
\end{tabular}
}
\caption{Leave-one-group-out ablation for the standalone LingRF model with the extended stylometric feature set. Values are test macro $F_1$ drops.}
\label{tab:lingrf_style_ablation_drops}
\end{table}

For the standalone LingRF model (Table~\ref{tab:lingrf_style_ablation_drops}), lexical diversity produces the largest and most consistent degradation across all evaluation settings. Functional markers and readability also contribute in some settings, especially on S1-en. In contrast, sentence structure, repetition, and punctuation show less stable behavior, including negative values on some datasets.

\begin{table}[!t]
\centering
\small
\setlength{\tabcolsep}{4pt}
\resizebox{\linewidth}{!}{%
\begin{tabular}{llrrrrr}
\hline
\textbf{Setting} & \textbf{Removed group} & \textbf{S1-en} & \textbf{S1-es} & \textbf{S2-en} & \textbf{S2-es} & \textbf{Avg.} \\
\hline

\multirow{7}{*}{Old-prob}
& Lexical diversity & \negdrop{-0.007} & \posdrop{+0.026} & \posdrop{+0.016} & \posdrop{+0.028} & \posdrop{+0.016} \\
& Sentence structure & +0.000 & \negdrop{-0.003} & \posdrop{+0.004} & \posdrop{+0.009} & \posdrop{+0.003} \\
& Repetition & \negdrop{-0.003} & \posdrop{+0.001} & \posdrop{+0.004} & \posdrop{+0.005} & \posdrop{+0.002} \\
& Word-level stats & \negdrop{-0.001} & \posdrop{+0.002} & \posdrop{+0.001} & \posdrop{+0.006} & \posdrop{+0.002} \\
& Functional markers & \negdrop{-0.001} & \posdrop{+0.001} & \negdrop{-0.001} & \posdrop{+0.006} & \posdrop{+0.001} \\
& Readability & \negdrop{-0.001} & \negdrop{-0.002} & +0.000 & \posdrop{+0.005} & +0.000 \\
& Punctuation & \negdrop{-0.001} & +0.000 & \posdrop{+0.006} & \posdrop{+0.003} & \posdrop{+0.002} \\

\hline

\multirow{7}{*}{Multilingual}
& Lexical diversity & \posdrop{+0.001} & \posdrop{+0.006} & \posdrop{+0.018} & \posdrop{+0.023} & \posdrop{+0.012} \\
& Sentence structure & +0.000 & \negdrop{-0.001} & \negdrop{-0.001} & \negdrop{-0.001} & \negdrop{-0.001} \\
& Repetition & +0.000 & \posdrop{+0.001} & \posdrop{+0.002} & \negdrop{-0.001} & +0.000 \\
& Word-level stats & \negdrop{-0.001} & \posdrop{+0.002} & \negdrop{-0.001} & \negdrop{-0.003} & \negdrop{-0.001} \\
& Functional markers & +0.000 & \negdrop{-0.002} & \posdrop{+0.001} & \negdrop{-0.002} & \negdrop{-0.001} \\
& Readability & \posdrop{+0.001} & \negdrop{-0.001} & +0.000 & \posdrop{+0.001} & +0.000 \\
& Punctuation & +0.000 & +0.000 & \negdrop{-0.001} & \negdrop{-0.003} & \negdrop{-0.001} \\

\hline
\end{tabular}
}
\caption{Leave-one-group-out ablation for LingRF + PredOut models with the extended stylometric feature set. Values are test macro $F_1$ drops.}
\label{tab:predout_style_ablation_drops}
\end{table}

In the PredOut settings (Table~\ref{tab:predout_style_ablation_drops}), the contribution of individual stylometric groups becomes smaller. Lexical diversity remains the strongest added stylometric group, with positive average drops in both Old-prob and Multilingual settings. Most other groups produce near-zero or inconsistent changes, and several groups have negative average values in the multilingual setting. This suggests that once probabilistic PredOut features are available, the added stylometric features provide only complementary signal, mainly through lexical diversity.

\textbf{Grouped permutation importance}. Table~\ref{tab:predout_multilingual_permutation} reports grouped permutation importance for the final LingRF + PredOut multilingual model with the extended stylometric set. The PredOut probabilities dominate the final classifier, but lexical diversity remains the strongest added non-probabilistic stylometric group.

\begin{table}[!t]
\centering
\small
\setlength{\tabcolsep}{3pt}
\resizebox{\columnwidth}{!}{%
\begin{tabular}{lrrrrr}
\hline
\textbf{Permuted group} & \textbf{S1-en} & \textbf{S1-es} & \textbf{S2-en} & \textbf{S2-es} & \textbf{Avg.} \\
\hline
PredOut probabilities & \posdrop{+0.407} & \posdrop{+0.343} & \posdrop{+0.269} & \posdrop{+0.178} & \posdrop{+0.299} \\
Original ling. features & \posdrop{+0.009} & \posdrop{+0.004} & \posdrop{+0.027} & \posdrop{+0.022} & \posdrop{+0.016} \\
Lexical diversity & \posdrop{+0.005} & \posdrop{+0.011} & \posdrop{+0.056} & \posdrop{+0.071} & \posdrop{+0.035} \\
Sentence structure & +0.000 & \negdrop{-0.002} & \posdrop{+0.001} & \posdrop{+0.001} & +0.000 \\
Repetition & +0.000 & \posdrop{+0.001} & \posdrop{+0.003} & +0.000 & \posdrop{+0.001} \\
Word-level stats & \negdrop{-0.001} & \negdrop{-0.001} & \posdrop{+0.002} & \posdrop{+0.003} & \posdrop{+0.001} \\
Functional markers & +0.000 & +0.000 & \posdrop{+0.001} & \posdrop{+0.001} & +0.000 \\
Readability & +0.000 & \negdrop{-0.001} & \posdrop{+0.002} & \posdrop{+0.002} & \posdrop{+0.001} \\
Punctuation & +0.000 & +0.000 & \posdrop{+0.001} & \posdrop{+0.003} & \posdrop{+0.001} \\
\hline
\end{tabular}
}
\caption{Grouped permutation importance for the multilingual LingRF + PredOut model. Values are test macro $F_1$ drops.}
\label{tab:predout_multilingual_permutation}
\end{table}

Grouped permutation importance confirms that the final LingRF + PredOut model is driven mainly by probabilistic PredOut features. The added stylometric groups play a secondary role, but lexical diversity is the only one that remains consistently informative after PredOut features are included. The other stylometric groups have near-zero effects, suggesting that their information is either redundant with stronger feature groups or too unstable to contribute reliably in the combined model.


\section{SHAP Interpretability}
\label{sec:app_shap}

We use SHAP to examine whether the feature groups identified by ablation also appear in local model explanations. Mean absolute SHAP values are used only as an importance measure. To estimate directional effects, we use signed SHAP values, since importance alone does not show whether high or low feature values push the model toward a class.

For each selected feature, we compute the Pearson correlation $r(x,\phi)$ between the feature value $x$ and its signed SHAP contribution $\phi$ on the test set. Positive values mean that larger feature values tend to push predictions toward the target class; negative values mean that smaller values do so. Table~\ref{tab:shap_direction_evidence} reports representative patterns aggregated across the evaluated language settings unless a pattern was observed only in one language.

\begin{table}[!t]
\centering
\small
\setlength{\tabcolsep}{3pt}
\resizebox{\columnwidth}{!}{%
\begin{tabular}{lllrr}
\hline
\textbf{Task} & \textbf{Signal} & \textbf{Target} & \textbf{Mean $|\phi|$} & \textbf{$r$} \\
\hline

\multicolumn{5}{l}{\textbf{LingRF (Extended)}} \\
\hline

S1 & \texttt{root\_ttr} & AI & 0.031 & +0.483 \\
S1 & lexical diversity features & AI & 0.019 & +0.267 \\
S1 & \texttt{flesch\_kincaid\_grade} & AI & 0.016 & +0.768 \\
S1 & \texttt{flesch\_reading\_ease} & AI & 0.011 & -0.791 \\

S2 & \texttt{root\_ttr} & A--C & 0.027 & +0.703 \\
S2 & \texttt{root\_ttr} & D--F & 0.026 & -0.816 \\

\hline

\multicolumn{5}{l}{\textbf{LingRF + PredOut (Multilingual)}} \\
\hline

S1 & \texttt{PRED\_PROB\_1} & AI & 0.145 & +0.988 \\
S1 & \texttt{root\_ttr} & AI & 0.018 & +0.443 \\
S2 & matching \texttt{PRED\_PROB\_i} & A--F & 0.037 & +0.961 \\

\hline
\end{tabular}
}
\caption{Representative signed-SHAP evidence.}
\label{tab:shap_direction_evidence}
\end{table}

The signed-SHAP results are consistent with the ablation analysis but also show that lexical diversity is not a universal marker of machine-generated text. In Subtask~1, several lexical diversity features tend to push the model toward the AI class. In Subtask~2, the same feature separates groups of generators in different directions. In the multilingual LingRF + PredOut model, probability features dominate the explanations, while lexical diversity remains a secondary stylometric signal.

We complement the signed-SHAP table with grouped mean absolute SHAP plots. These plots are not used for directional claims; they only show which feature groups receive the largest average SHAP contributions. We report representative Spanish results for both subtasks because they capture the main trends observed across languages.

\begin{figure*}[!t]
\centering

\begin{subfigure}{\textwidth}
\centering
\includegraphics[width=0.98\linewidth, trim={0 0 0 1.3cm}, clip]{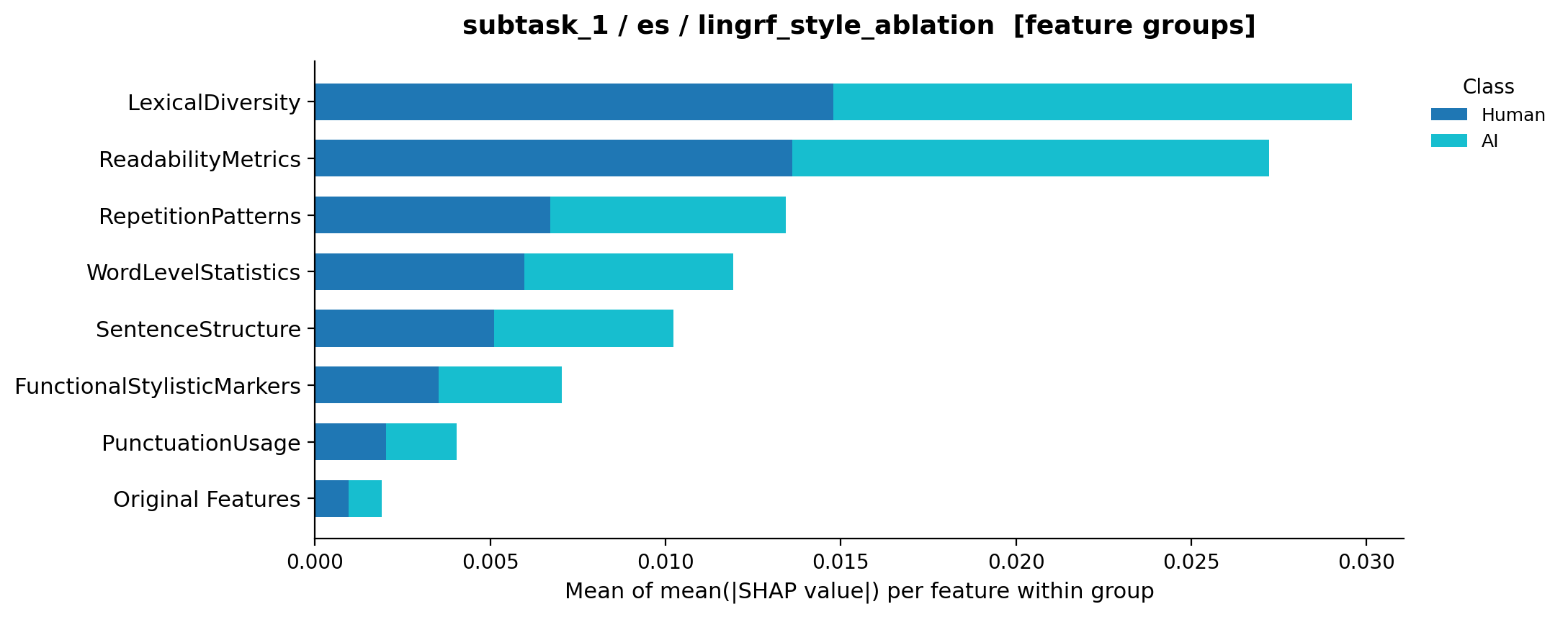}
\caption{Subtask 1}
\end{subfigure}

\vspace{2mm}

\begin{subfigure}{\textwidth}
\centering
\includegraphics[width=0.98\linewidth, trim={0 0 0 1.3cm}, clip]{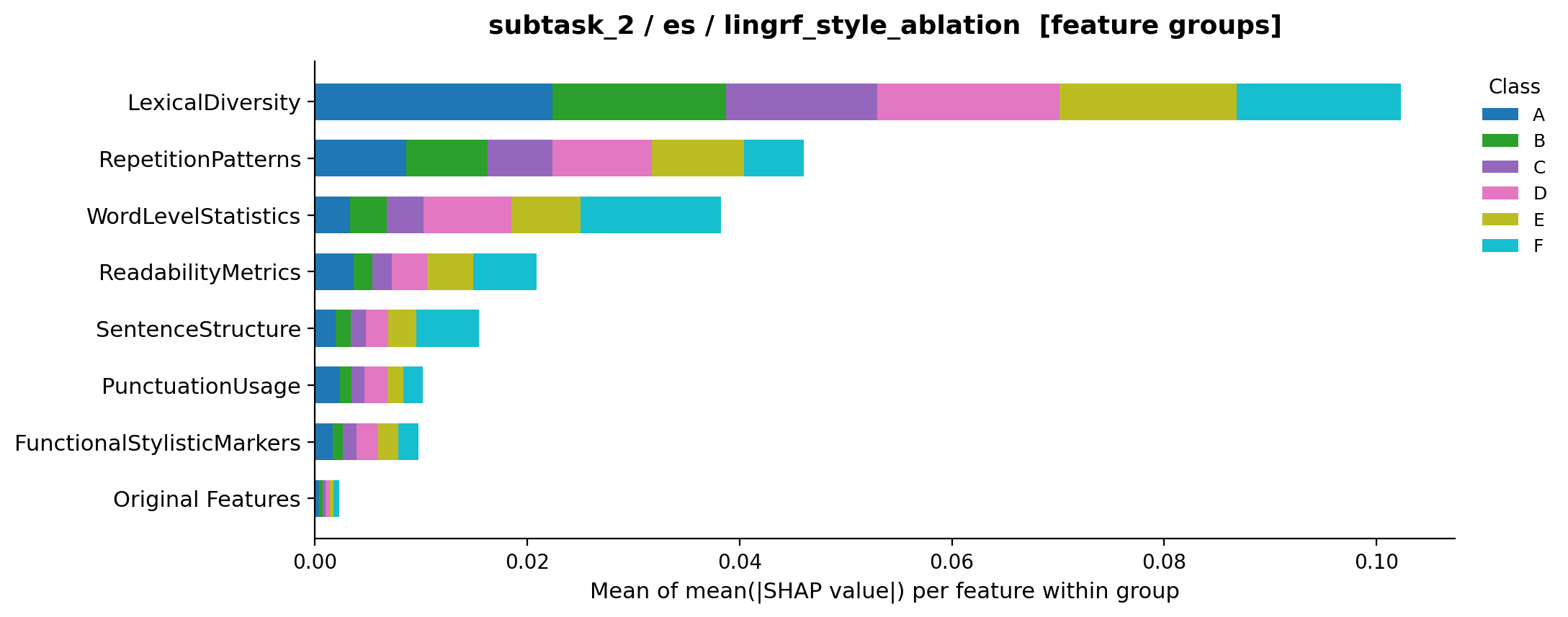}
\caption{Subtask 2}
\end{subfigure}

\caption{
Grouped mean absolute SHAP values for the standalone LingRF model with the extended stylometric feature set.
}
\label{fig:shap_lingrf_grouped}
\end{figure*}

\begin{figure*}[!t]
\centering

\begin{subfigure}{\textwidth}
\centering
\includegraphics[width=0.98\linewidth, trim={0 0 0 1.3cm}, clip]{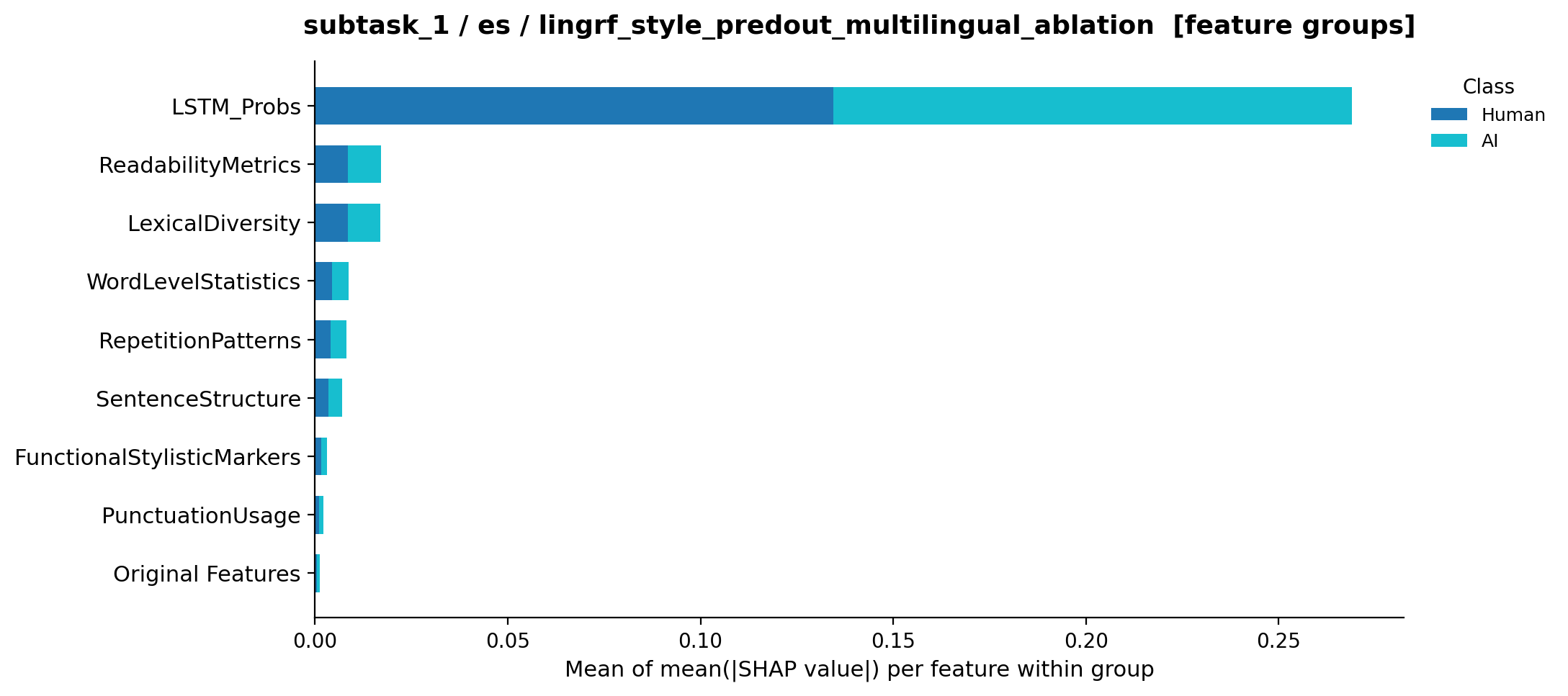}
\caption{Subtask 1}
\end{subfigure}

\vspace{2mm}

\begin{subfigure}{\textwidth}
\centering
\includegraphics[width=0.98\linewidth, trim={0 0 0 1.3cm}, clip]{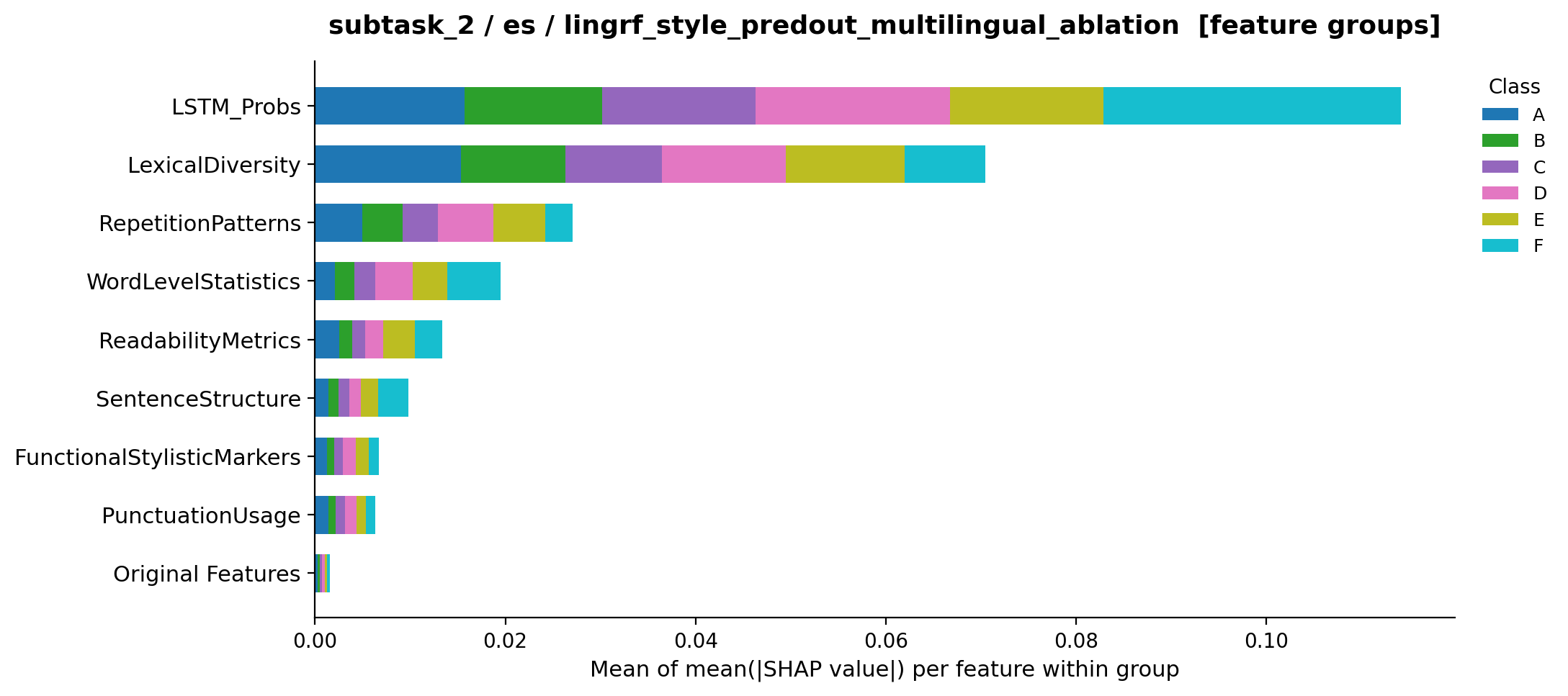}
\caption{Subtask 2}
\end{subfigure}

\caption{Grouped mean absolute SHAP values for the multilingual LingRF + PredOut model with the extended stylometric feature set.}
\label{fig:shap_predout_grouped}
\end{figure*}

The grouped SHAP plots are consistent with the ablation and permutation analyses. In the standalone LingRF model, stylometric features provide most of the explanatory signal, with lexical diversity being the strongest group, especially in Subtask~2. After adding PredOut probabilities, the explanations shift toward the probabilistic features, while lexical diversity remains the strongest secondary stylometric signal.

\end{document}